%% file: acl_latex.tex
\pdfoutput=1

\documentclass[11pt]{article}

\usepackage{acl}

\usepackage{times}
\usepackage{latexsym}
\usepackage[most]{tcolorbox}
\usepackage{multicol}
\usepackage{booktabs}
\usepackage{longtable}
\usepackage{outlines}
\usepackage{enumitem}

\usepackage{hyperref} 
\usepackage{subcaption}

\tcbset{
  myguidelinesbox/.style={
    colback=blue!5!white,
    colframe=blue!75!black,
    fonttitle=\bfseries,
    title=Project Guidelines,
    sharp corners,
    boxrule=0.8pt,
    width=\textwidth,
    enhanced,
  }
}

\usepackage[T1]{fontenc}

\usepackage[utf8]{inputenc}

\usepackage{microtype}

\usepackage{inconsolata}

\usepackage{graphicx}

\title{NLP for Social Good: A Survey and Outlook of \\ Challenges, Opportunities, and Responsible Deployment}

\author{
Antonia Karamolegkou\textsuperscript{1},
\\
\bf Angana Borah\textsuperscript{2},
 Eunjung Cho\textsuperscript{3},
 Sagnik Ray Choudhury\textsuperscript{4},
 Martina Galletti\textsuperscript{5,6},
Pranav Gupta\textsuperscript{7},
\\
\bf
Oana Ignat\textsuperscript{8},
Priyanka Kargupta\textsuperscript{9},
Neema Kotonya\textsuperscript{10},
Hemank Lamba\textsuperscript{10},
Sun-Joo Lee\textsuperscript{11},
Arushi Mangla\textsuperscript{8},
\\
\bf
Ishani Mondal\textsuperscript{12},
Fatima Zahra Moudakir\textsuperscript{13,14,15},
Deniz Nazar\textsuperscript{16},
Poli Nemkova\textsuperscript{4,11},
Dina Pisarevskaya\textsuperscript{17},
\\
\bf
Naquee Rizwan\textsuperscript{18},
Nazanin Sabri\textsuperscript{19},
Keenan Samway\textsuperscript{13},
Dominik Stammbach\textsuperscript{20},
Anna Steinberg Schulten\textsuperscript{21,22},
\\
\bf   
David Tomás\textsuperscript{23},
Steven R Wilson\textsuperscript{24},
Bowen Yi\textsuperscript{2,25},
Jessica Zhu\textsuperscript{12},
Arkaitz Zubiaga\textsuperscript{17},
Anders Søgaard\textsuperscript{1},
\\
\bf
Alexander Fraser\textsuperscript{22,26},
Zhijing Jin\textsuperscript{3,13,14,15},
Rada Mihalcea\textsuperscript{2},
Joel R. Tetreault\textsuperscript{10},
Daryna Dementieva\textsuperscript{22,26}
\\
\\
\footnotesize \textsuperscript{1}University of Copenhagen
\textsuperscript{2}University of Michigan-Ann Arbor
\textsuperscript{3}ETH Zurich
\textsuperscript{4}University of North Texas
\textsuperscript{5}Sony Computer Science Laboratories - Paris
\\
\footnotesize
\textsuperscript{6}University of Rome ``La Sapienza''
\textsuperscript{7}Lowe's
\textsuperscript{8}Santa Clara University
\textsuperscript{9}University of Illinois Urbana-Champaign
\textsuperscript{10}Dataminr
\\
\footnotesize
\textsuperscript{11}United Nations Development Programme (UNDP)
\textsuperscript{12}University of Maryland, College Park
\textsuperscript{13}Max Planck Institute for Intelligent Systems, Tübingen
\\
\footnotesize
\textsuperscript{14}Vector Institute
\textsuperscript{15}University of Toronto
\textsuperscript{16}University of Washington
\textsuperscript{17}Queen Mary University of London
\textsuperscript{18}IIT Kharagpur
\\
\footnotesize
\textsuperscript{19}University of California San Diego
\textsuperscript{20}Princeton University
\textsuperscript{21}LMU Munich
\textsuperscript{22}Munich Center for Machine Learning (MCML)
\textsuperscript{23}University of Alicante
\\
\footnotesize
\textsuperscript{24}University of Michigan-Flint
\textsuperscript{25}University of Southern California
\textsuperscript{26}Technical University of Munich
\\
\small
\textbf{Accepted to EACL 2026 (Main Conference), Rabat, Morocco.} \\
\small Correspondence: \href{mailto:antoniakrm16@gmail.com}{antoniakrm16@gmail.com}, \href{mailto:daryna.dementieva@tum.de}{daryna.dementieva@tum.de}
}

\input{latex/colourcodes}
\begin{document}
\maketitle
\begin{abstract}
Natural language processing (NLP) now shapes many aspects of our world, yet its potential for positive social impact is underexplored. This paper surveys work in ``NLP for Social Good" (NLP4SG) across nine domains relevant to global development and risk agendas, summarizing principal tasks and challenges. We  analyze ACL Anthology trends, finding that inclusion and AI harms attract the most research, while domains such as poverty, peacebuilding, and environmental protection remain underexplored. Guided by our review, we outline opportunities for responsible and equitable NLP and conclude with a call for cross-disciplinary partnerships and human-centered approaches to ensure that future NLP technologies advance the public good.
\end{abstract}

\input{latex/1_INTRO}

\input{latex/2_TOPICS}

\input{latex/3_summary}

\section*{Limitations}
This paper offers a high-level interdisciplinary perspective on aligning NLP research with societal needs, grounded in the UN Sustainable Development Goals (SDGs) and the World Economic Forum’s Global Risks Report. While our proposed framework maps NLP research directions to these agendas, the related works list discussed is not exhaustive. While we aimed to cover the most impactful topics based on the authors’ expertise, we acknowledge that ongoing advancements in NLP may enable new tasks and uncover deeper layers of impact beyond what could be envisioned at the time of writing.

Furthermore, while we highlight areas of overlap between the SDGs and global risks, the mappings remain somewhat subjective. Another key assumption we made is that positive impact is highly aligned with the UN Sustainable Development Goals - but this might not be true i.e. positive impact could still be derived from NLP tools without there being an explicit alignment towards SDGs. At the same time, we believe that our work will open more cross-disciplinary discussions and more ideas for NLP4SG applications.

\section*{Ethics Statement}
This work is grounded in the belief that NLP research should be aligned with broader societal priorities and developed with care for its downstream impact. All ACL materials used for our statistical analysis are licensed using the Creative Commons 3.0 BY-NC-SA. In proposing mappings between NLP directions, global goals, and risks, we are mindful of the potential for unintended consequences. The open areas of work mentioned throughout the paper are just suggestive, and the practitioners, when working on them, should carefully evaluate the setting and downstream impacts.

We emphasize the importance of interdisciplinary collaboration and the inclusion of marginalized perspectives in shaping responsible NLP research. Where possible, we cite and build on existing initiatives in NLP for social good, and we support efforts to foreground equity, inclusivity, and transparency in both research framing and methodology. Throughout this paper, we emphasize that NLP systems should augment rather than replace human expertise, particularly in sensitive or high-stakes contexts. Finally, we acknowledge that AI assistants were used for proofreading this paper, a practice we consider ethically acceptable under appropriate usage.

\section*{Acknowledgments}
\input{latex/4_Acknowledgments}

\input{foo.bbl}
\onecolumn
\appendix

\section{Quantitative Analysis}
\label{sec:appendix-llm-annotation}
\subsection{Data Preparation}

The \textit{PaperAnalyzer} framework \citep{adauto-etal-2023-beyond} introduced a large-scale methodology for classifying research papers into socially relevant categories, drawing connections between NLP tasks, methods, and broader social goals such as the United Nations Sustainable Development Goals (SDGs). By leveraging large language models (LLMs) for annotation, \textit{PaperAnalyzer} demonstrated how automated approaches can reduce the dependence on costly manual labeling while enabling scalable analysis of research trends.  

Our work is inspired by this methodology and adapts it to a more focused investigation of yearly trends within the ACL Anthology. Specifically, we designed an annotation pipeline to assign papers to the set of key NLP application areas, enabling us to capture how the community’s research priorities have evolved over time. 

For this study, we focused on ACL Anthology papers published between 2019 and 2024, including both main conference and workshop proceedings. This filtering yielded 47,078 papers out of the 113,207 entries in the Anthology. Across this subset, approximately 7\% of papers lack abstracts; these papers were retained, with only their titles used for annotation. 

\subsection{Annotation of Key NLP Domains}

\subsubsection{Methodology}
To analyze the research landscape from a social good perspective, we use a zero-shot annotation pipeline that categorized papers into nine NLP application areas: \textit{Healthcare, Education, Poverty, Peacebuilding, Environment Protection, Inclusion \& Inequalities, Online Harms, Misinformation, and AI Harms}. Papers could be assigned to multiple domains when appropriate, while those not fitting any category were labeled as \textit{Unrelated}.

Each paper was evaluated using its title and abstract (when available). The model was explicitly instructed with the full list of domains, together with concise descriptions for each. These descriptions were initially drafted with the assistance of \texttt{ChatGPT}, but were subsequently reviewed and edited by the authors to ensure contextual appropriateness for the study. Our prompt can be viewed in Figure \ref{fig:annotation_prompt}.

For the large-scale annotation of the full dataset, we used OpenAI's GPT-4.1 mini (\texttt{openai/gpt-4.1-mini-2025-04-14}), which provided both efficiency and value at scale. All queries were executed with \texttt{temperature = 0} to ensure deterministic and reproducible outputs. The pipeline was applied to the complete set of 47,078 ACL Anthology papers published between 2019 and 2024. The total cost to annotate all papers was approximately \$12 USD.

\subsubsection{Results}

Overall, the pipeline annotated 36,030 papers (approximately 76.5\%) with at least one valid domain label, with each of these papers receiving an average of 1.82 domain assignments. During annotation, 109 papers received at least one invalid annotation. Of these, 107 papers contained a mix of valid and invalid annotations, which we handled by retaining only the valid domain labels. The remaining 2 papers contained exclusively invalid annotations and were therefore treated as \textit{Unrelated}.

\begin{figure*}[ht!]
    \centering
    \begin{subfigure}[t]{0.49\linewidth}
        \centering
        \includegraphics[width=\linewidth]{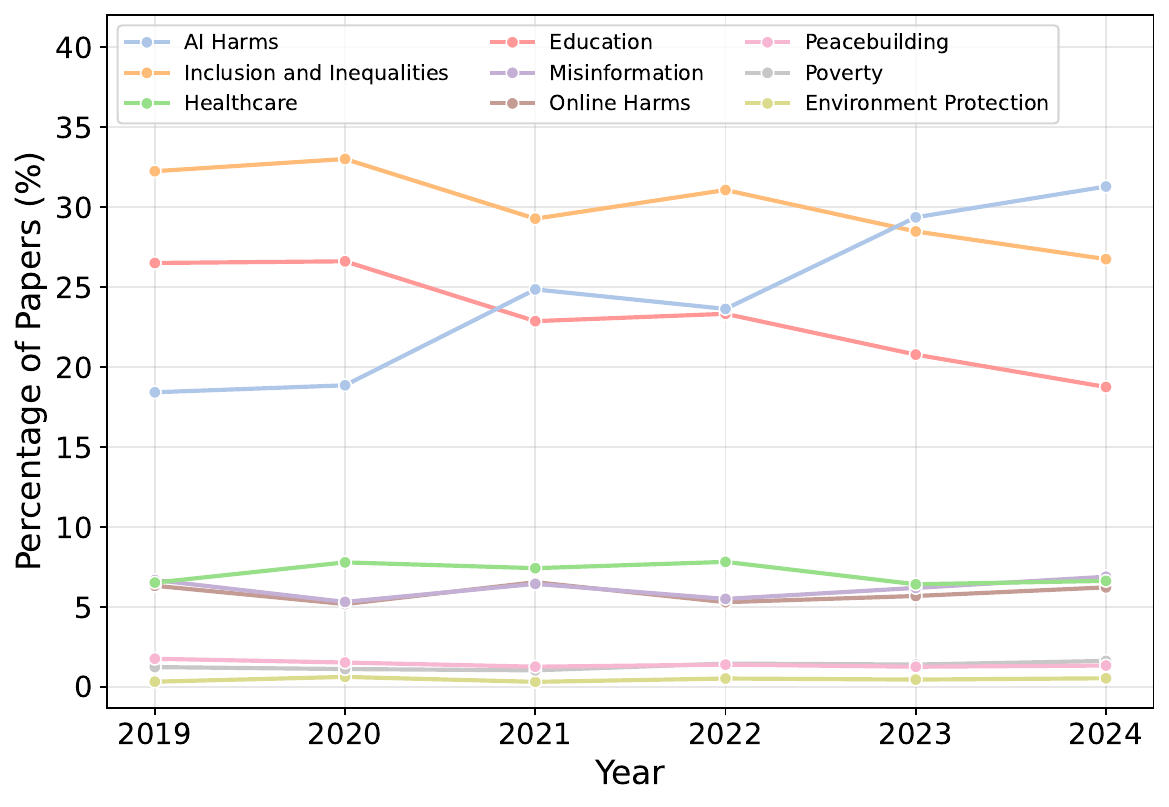}
        \caption{Conference Papers}
        \label{fig:paper_trends_conference_normalized}
    \end{subfigure}
    \hfill
    \begin{subfigure}[t]{0.49\linewidth}
        \centering
        \includegraphics[width=\linewidth]{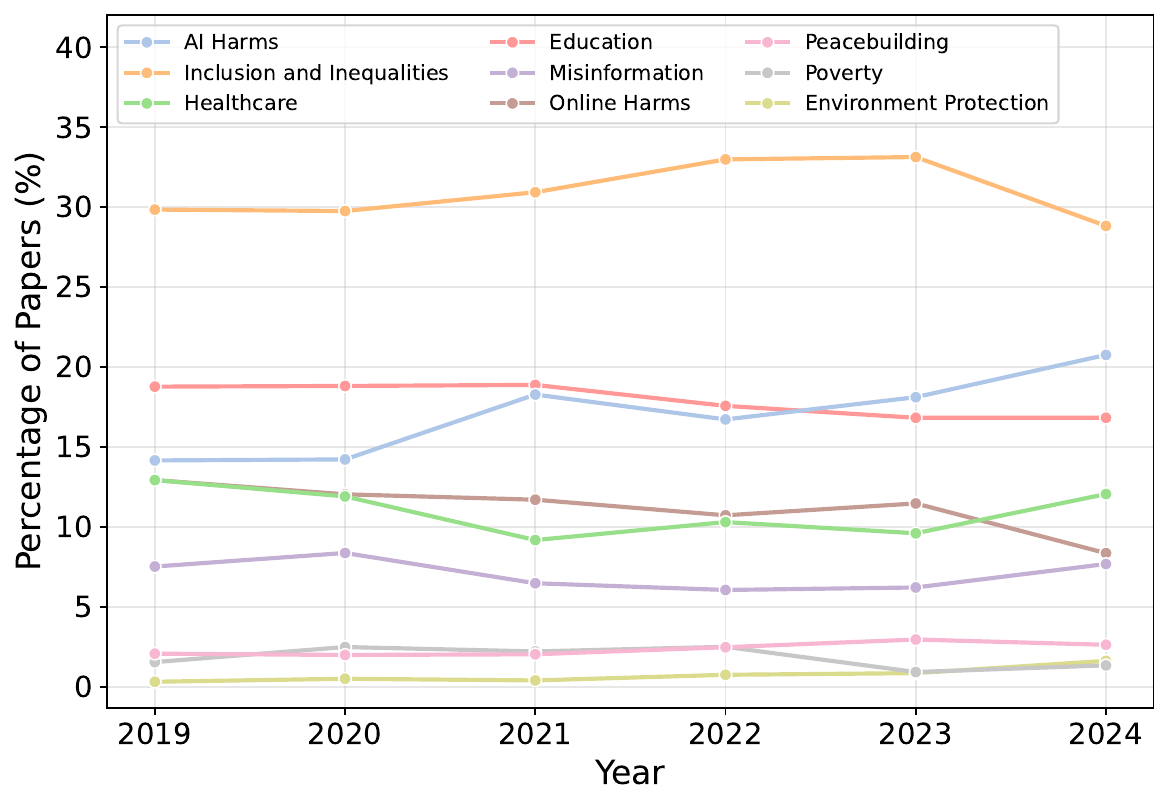}
        \caption{Workshop Papers}
        \label{fig:paper_trends_workshop_normalized}
    \end{subfigure}
    \caption{Overview of the normalized NLP domain distributions for conference papers ($n=32,886$) and workshop papers ($n=12,723$) from 2019 to 2024.}
    \label{fig:paper_trends_combined}
\end{figure*}

To examine how social good research is distributed across different publication venues, we analyzed the distribution of the key NLP domains across conference papers and workshop papers separately, as shown in Figure \ref{fig:paper_trends_combined}. Conferences were identified by filtering for entries where ``Conference,'' ``Findings,'' or ``Annual Meeting'' appeared in the booktitle field, yielding 32,886 papers. Workshop papers were identified by filtering for entries where ``Workshop'' appeared in the booktitle field, yielding 12,723 papers.

\begin{figure*}[ht!]
    \centering
    \includegraphics[width=0.6\linewidth]{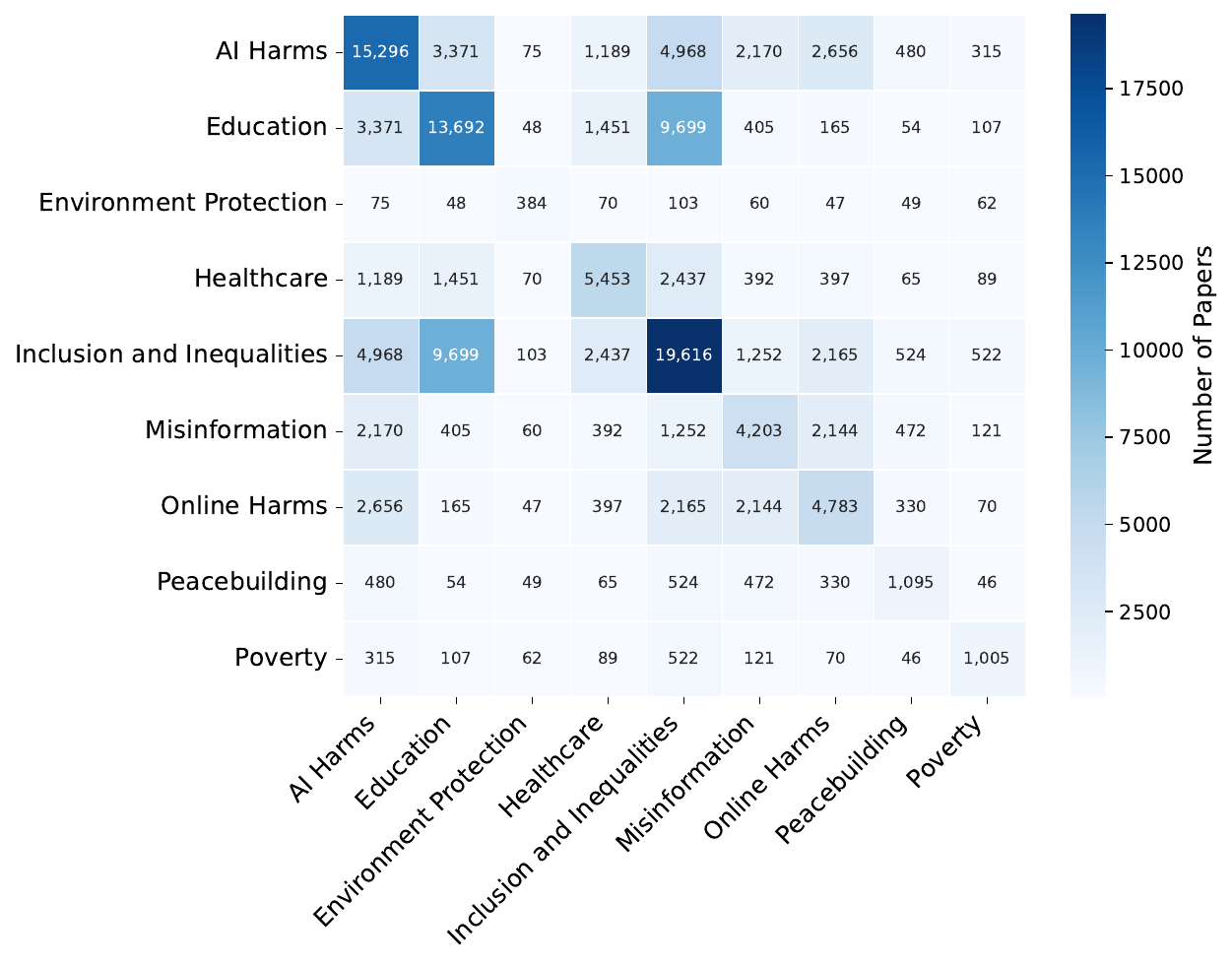}
    \caption{Co-occurrence heatmap of annotated NLP domains.}
    \label{fig:paper_cooccurrence_heatmap}
\end{figure*}

Based on these graphs, workshops show more pronounced year-to-year fluctuations and a stronger emphasis on emerging or experimental areas. While AI Harms and Inclusion \& Inequalities remain dominant in both, workshop papers show relatively higher proportions of Online Harms and Misinformation, suggesting that exploratory and early-stage research on socially sensitive topics tends to appear more frequently in workshop venues before reaching conference-level maturity.

Overall, the co-occurrence heatmap in Figure \ref{fig:paper_cooccurrence_heatmap} confirms these broader patterns, showing that research on \textit{AI Harms} and \textit{Inclusion \& Inequalities} often overlaps with multiple domains, making them central themes in NLP4SG. By contrast, areas such as \textit{Poverty} and \textit{Peacebuilding} appear more isolated, suggesting that while ethical and fairness-related work has gained strong interdisciplinary traction, other social domains remain comparatively underexplored.

\begin{figure*}[h!]
\begin{tcolorbox}[myguidelinesbox, title=LLM Annotation Prompt]

\begin{footnotesize}
    
You will be given the title and abstract of a research paper.

Your task is to annotate the paper with the relevant social impact domains.
\begin{itemize}
  \renewcommand{\labelitemi}{-}
    \item Review the list of domains provided.
    \item Select all domains that are relevant to the paper based on its title and abstract; you can select multiple.
    \item If none of the domains apply, respond with "Unrelated".
    \item Do not explain your choices.
    \item Do not include anything other than the list of exact domain names or the word "Unrelated".
\end{itemize}

**Domains**:

Healthcare – Enhance physical and mental well-being by improving access, diagnosis, and support through language technologies.

Education – Ensure inclusive and equitable quality education with NLP tools for learning, teaching, and accessibility.

Poverty – Reduce economic hardship by enabling data-driven policies, resource distribution, and financial inclusion with NLP.

Peacebuilding – Promote peaceful societies by using NLP for dynamic conflict analysis, human rights protection, and crisis response through early-warning and mediation support.

Environment Protection – Address environmental risks by monitoring and analyzing them, and supporting sustainability, climate action and environmental responsibility using NLP techniques.

Inclusion and Inequalities – Foster fairness and representation by using NLP to address bias, improve accessibility, and empower marginalized voices.

Online Harms – Make digital spaces safe by detecting and mitigating hate speech, harassment, and harmful online behaviors.

Misinformation – Safeguard information integrity by detecting, verifying, and countering false and misleading content.

AI Harms – Build responsible AI by improving transparency, accountability, and fairness to prevent unintended harm from NLP systems.
\\\\
**Paper Title**:

\{paper\_title\}
\\\\
**Paper Abstract**:

\{paper\_abstract\}
\\\\
Please respond in the following format only:

Domain; Domain; ...   OR   Unrelated
       
\end{footnotesize}

\end{tcolorbox}
\caption{Prompt used to annotate ACL Anthology papers with the key NLP research directions.}
\label{fig:annotation_prompt}
\end{figure*}

\subsection{Annotation of NLP Tasks and Methods}
Building on the approach proposed in Task 3 of \textit{PaperAnalyzer} \citep{adauto-etal-2023-beyond}, which analyzed NLP research for social good through its underlying tasks and methods, we aimed to reproduce a similar insight. The goal was to capture how specific techniques and problem types evolve across socially relevant research areas, complementing our domain-level annotation.

\subsubsection{Methodology}
We began with the set of tasks and methods previously generated from our 900-paper sample subset. We prompted Gemma 3 (\texttt{google/gemma-3-27b-it}) to identify and extract the tasks and methods mentioned in each paper’s title and abstract. This process yielded approximately 750 unfiltered task terms and nearly 1,000 unfiltered method terms. To consolidate these into a more structured taxonomy, we provided all extracted terms to ChatGPT and asked it to group them into canonical categories representing the most common NLP tasks and methods. We then manually reviewed each category to ensure correctness. 

The resulting taxonomy included 45 distinct task categories and 49 method categories. These categories were then used to annotate the full set dataset, where we used Google's Gemini 2.5 Flash (\texttt{google/gemini-2.5-flash}) to assign one or more task and method labels to each paper based on its title and abstract. The prompt used for this extraction (shown in Figure \ref{fig:annotation_prompt_tasks_methods}) instructed the model to return only relevant tasks and methods in a standardized format, ensuring consistency across the dataset. The total cost to annotate all papers was approximately \$22 USD.

\subsubsection{Results}
\begin{figure*}[ht!]
    \centering
    \includegraphics[width=\linewidth]{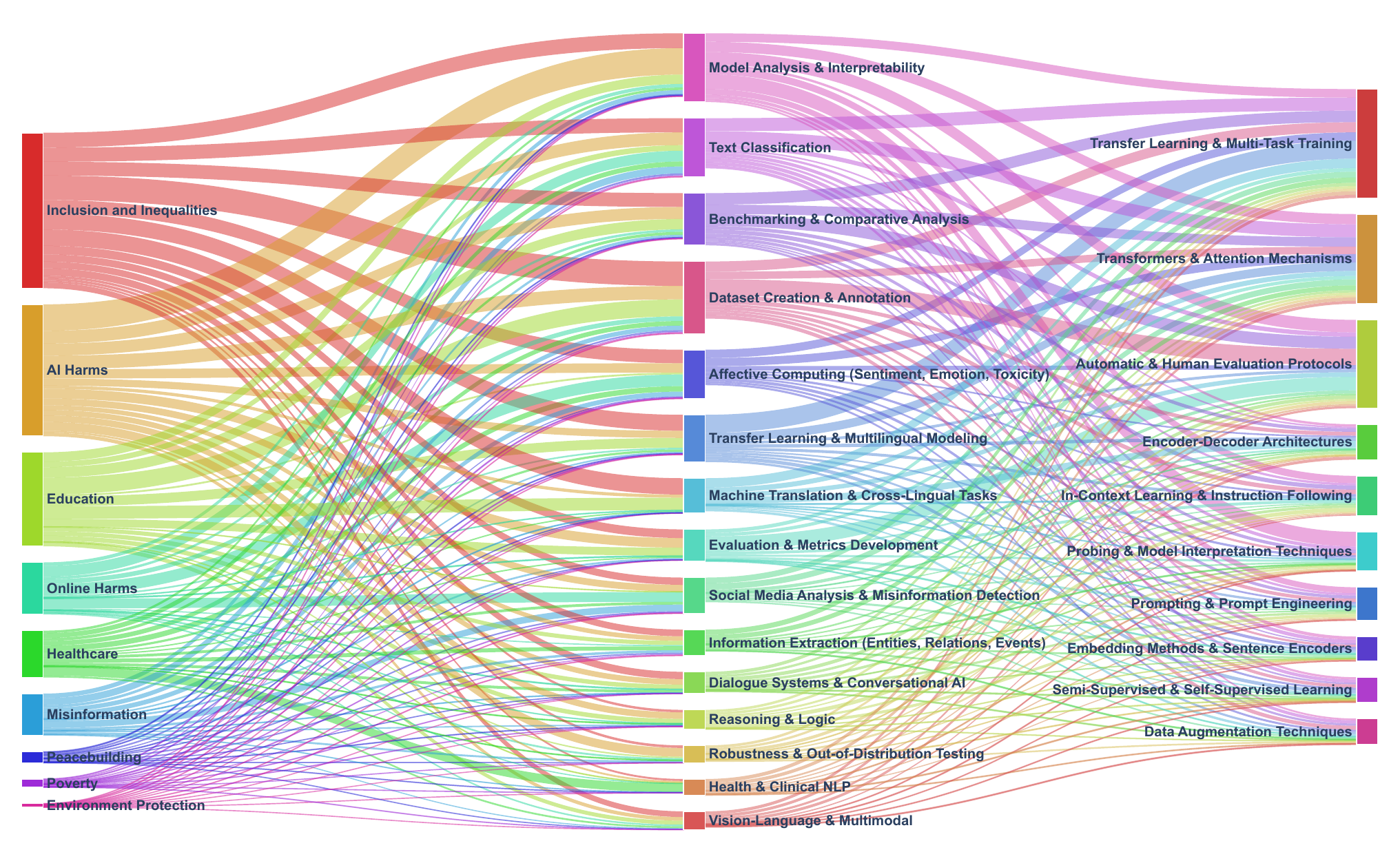}
    \caption{Diagram showing the flow of research focus from the key NLP domains to the top 15 NLP tasks and the corresponding top 10 methodological approaches.}
    \label{fig:paper_sankey}
\end{figure*}

After the annotation process, only one paper contained an invalid task label, which was manually removed. Sixteen papers contained no valid method annotations; for these, the method field was replaced with the label \textit{Unknown}. Overall, task coverage was nearly complete, with 46{,}996 papers (99.83\%) successfully assigned one or more task labels and an average of 2.74 tasks per paper. Method coverage was similarly high, with 45{,}034 papers (95.66\%) receiving at least one method label and an average of 2.22 methods per paper. Finally, to visualize the relationships between the annotated categories, we generated a Sankey diagram (Figure \ref{fig:paper_sankey}) linking the domains, tasks, and methods, illustrating how research themes and techniques intersect across the ACL Anthology corpus.

The diagram reveals that recent NLP4SG research has increasingly centered on \textit{AI Harms} and \textit{Inclusion \& Inequalities}, highlighting a shift toward fairness, bias mitigation, and ethical analysis of large language models. These domains are most frequently associated with \textit{Dataset Creation}, \textit{Model Analysis \& Interpretability} and \textit{Text Classification}, indicating a growing focus on building representative data, evaluating model behavior and developing practical applications for social impact.

\begin{figure*}[ht!]
    \centering
    \begin{subfigure}[t]{0.49\linewidth}
        \centering
        \includegraphics[width=\linewidth]{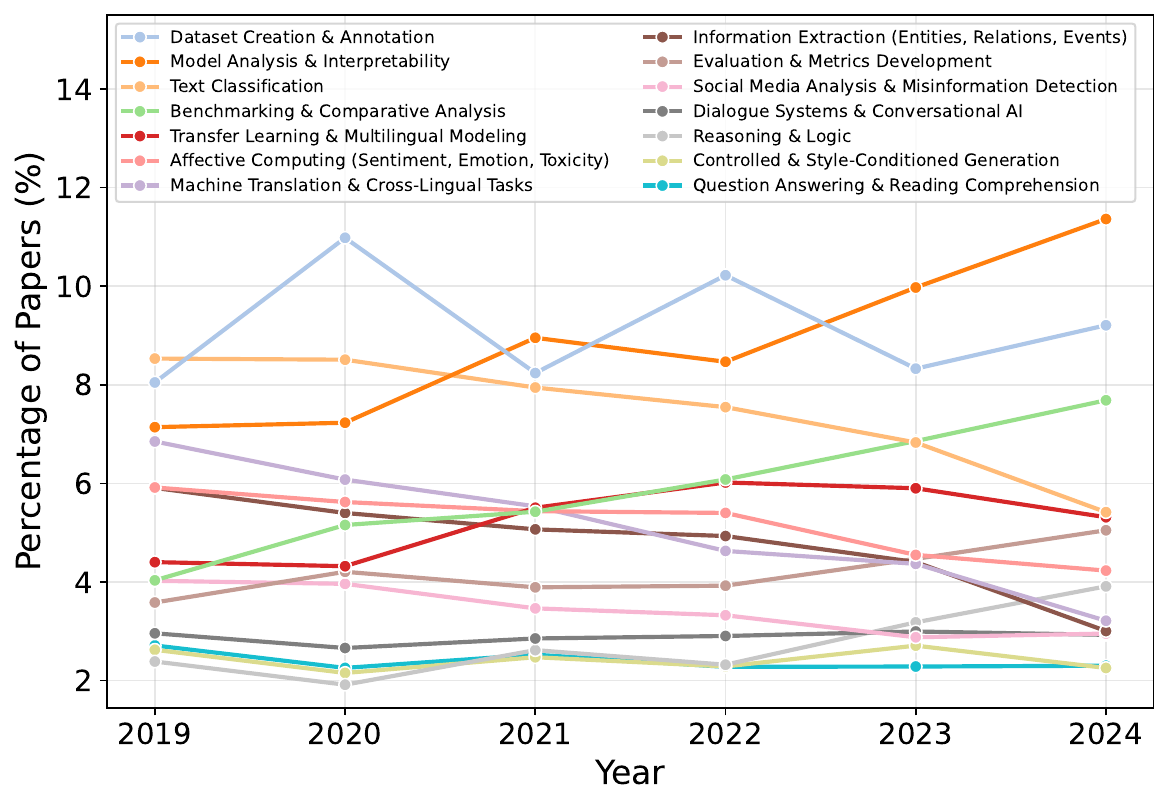}
        \caption{Tasks}
        \label{fig:paper_task_trends}
    \end{subfigure}
    \hfill
    \begin{subfigure}[t]{0.49\linewidth}
        \centering
        \includegraphics[width=\linewidth]{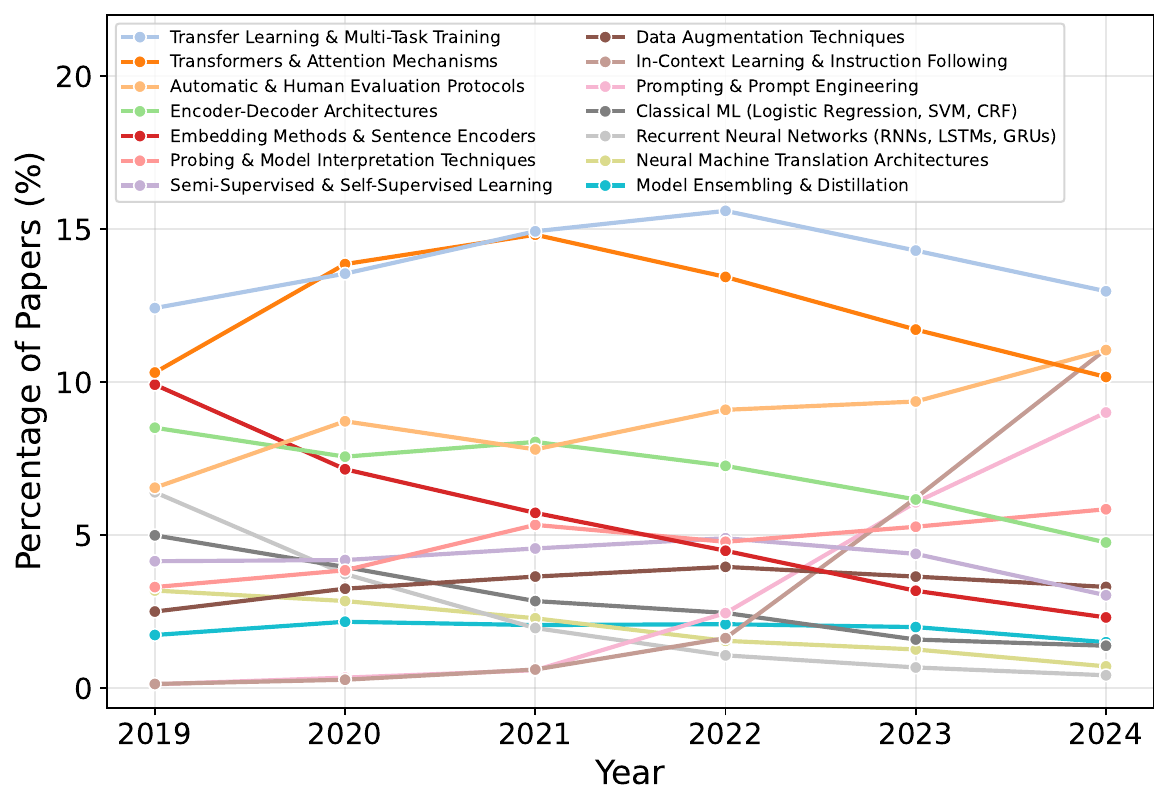}
        \caption{Methods}
        \label{fig:paper_method_trends}
    \end{subfigure}
    \caption{Overview of the normalized NLP task and method distributions for ACL Anthology papers from 2019 to 2024. Each plot shows trendlines for the 14 most frequent tasks and methods, selected based on their average frequency across all years.}
    \label{fig:task_and_method_trends}
\end{figure*}

Together, the task and method trends (Figure \ref{fig:task_and_method_trends}) illustrate a clear evolution in NLP4SG research. Tasks such as \textit{Model Analysis \& Interpretability}, \textit{Dataset Creation \& Annotation}, and \textit{Text Classification} have gained prominence, reflecting a move toward model evaluation and analysis. Meanwhile, \textit{Transformers} and \textit{Transfer Learning} are gradually giving way to approaches like \textit{Prompting} and \textit{In-Context Learning}, highlighting the growing integration of LLM-driven, instruction-based methodologies.

\begin{figure*}[h!]
\begin{tcolorbox}[myguidelinesbox, title=LLM Annotation Prompt]

\begin{footnotesize}
    
You will be given the title and abstract of a research paper.

Your task is to annotate the paper with the primary NLP task(s) and method(s) that this paper is dealing with.

\begin{itemize}
  \renewcommand{\labelitemi}{-}
    \item Review the list of tasks and methods provided.
    \item Select all tasks and methods that are relevant to the paper based on its title and abstract; you can select multiple.
    \item If none of the tasks or methods apply, respond with "Unknown".
    \item Do not explain your choices.
    \item Do not include anything other than the list of task or method labels or the word "Unknown".
    \item Make sure to respond with the exact label for the tasks and methods. Do not modify them in any way.
\end{itemize}

**Task List**:

T1. Text Classification\\
T2. Sequence Labeling\\
...\\
T44. Web Interaction \& Automation\\
T45. Benchmarking \& Comparative Analysis
\\\\
**Method List**:

M1. Recurrent Neural Networks (RNNs, LSTMs, GRUs)\\
M2. Transformers \& Attention Mechanisms\\
...\\
M48. Classical ML (Logistic Regression, SVM, CRF)\\
M49. Tool Integration \& API Augmentation
\\\\
**Paper Title**:

\{paper\_title\} 
\\\\
**Paper Abstract**:

\{paper\_abstract\} 
\\\\
Please respond in the following format **only**:

Tasks: T\#; T\#; ...

Methods: M\#; M\#; ...
       
\end{footnotesize}

\end{tcolorbox}
\caption{Prompt used to annotate ACL Anthology papers with NLP tasks and methods.}
\label{fig:annotation_prompt_tasks_methods}
\end{figure*}

\newpage

\section{Paper Selection: Thematic Tables of Datasets, NLP Tasks, Evaluation Metrics, and References}
\label{sec:appendix}

\subsection{Health and Well-being}

Mental health is a key component of human health, encompassing our emotional, psychological, and social well-being \cite{samhsa2023mental}. Currently, mental health issues are a multifactorial global crisis complicated by individual risk factors and various socioeconomic and clinical factors, but NLP methods show promising potential to enhance mental healthcare \cite{zhang2022natural}. In this context, we define NLP tasks as activities performed by NLP techniques in roles similar to counselors and clients. In the role of \textbf{counselors}, NLP tools engage in several core tasks: (1) \textbf{Detection and classification} of mental health conditions, such as depression \cite{giuntini2020review, mahdy2020comparative, khan2018analysis} and addiction \cite{Yang2023IdentifyingSO,Kwon2023ODDAB, Ni2021AutomatedDO}, using data sources like clinical notes \cite{Panaite2022TheVO, calvo2017natural} and social media posts \cite{skaik2020using, chancellor2020methods, rissola2021survey}; (2) \textbf{Responding} to users by interpreting their emotional states \cite{Warikoo2022NLPMP, Sabour2024EmoBenchET, Grandi2024TheES}, generating therapeutic and empathetic responses \cite{shen-etal-2020-counseling, Grandi2024TheES, sharma-etal-2023-cognitive, nazarova2023application, zhou2025crispcognitiverestructuringnegative}, and providing actionable feedback for support quality \cite{min-etal-2023-verve, chaszczewicz-etal-2024-multi, althoff-etal-2016-large}; (3) \textbf{Tracking} emotion and mood via time-series data analysis \cite{osi2024WarEM} and detecting mental health crises over time \cite{Gong2019MachineLD, yuan2023mental}. Conversely, when NLP tools serve as \textbf{clients}, they typically simulate client personas from diverse backgrounds to train counselors \cite{louie-etal-2024-roleplay, liu2025eeyorerealisticdepressionsimulation, 10.1145/3710993, kampman2025conversationalselfplaydiscoveringunderstanding}. The literature for this section was selected based on the existing survey papers such as \cite{zhang2022natural,Malgaroli2023NaturalLP} as well as keyword search (e.g., "AI for mental health", "LLM-based mental health applications", "mental health chatbots", "AI therapy").

Physical well-being refers to maintaining one's bodily health through behaviors such as regular physical activity, adequate sleep, balanced nutrition, good hygiene, and avoidance of harmful substances \cite{Lotan:2005}. To stimulate discussion on the role of NLP in physical well-being, the works included in the paper were selected based on a targeted keyword search including ``physical activity'', ``sleep'', ``nutrition'', ``hygiene'', ``substance use'', and ``clinical report analysis'', which also plays a central role in physical health. NLP techniques have been applied to various aspects of physical well-being, leveraging unstructured text data to monitor behaviors and inform interventions. For instance, physical activity and sedentary behavior can be tracked through analysis of social media using NLP-driven health surveillance systems that mine Twitter posts to estimate physical activity levels, sedentary behavior, and sleep patterns in populations \cite{Sakib:2021,Shakeri:2022}. For nutrition, NLP has been employed to assess dietary habits by using models that can automatically classify foods and meals from descriptions, even providing personalized diet advice \cite{Erp:2021,Hu:2023}. Regarding hygiene, during the COVID-19 pandemic NLP was used to estimate public perceptions of mask-wearing and other hygiene practices by mining social media posts \cite{Al-Garadi:2022}. Likewise, in harmful habit avoidance, NLP methods help identify substance use patterns and risky behaviors from online text \cite{Hu:2021,Lin:2023}. The analysis of clinical reports also plays a central role in physical health. These documents capture critical patient information that is often not represented in structured data fields. Consequently, clinical report analysis has emerged as a core subtask within NLP for physical health, enabling the extraction, classification, and summarization of medically relevant information directly from unstructured clinical narratives \cite{Landolsi:2023}. In this research area there is a growing role of LLMs in extracting information from clinical reports. The work by \cite{Mannhardt:2024} shows how GPT-4 can support patients by simplifying clinical notes, improving comprehension and confidence, though with some factual inaccuracies. In \cite{Guluzade:2025}, the authors introduce a large annotated dataset (ELMTEX) and find that fine-tuned small LLMs outperform larger ones in extracting structured information efficiently. These findings are confirmed in pathology reports, where fine-tuned models achieve higher accuracy and fewer hallucinations than prompt-based methods \cite{Park:2024}. LLMs can also be used in clinical reports to generate and summarize documentation such as patient notes, discharge summaries, and case reports, offering improvements in efficiency, organization, and standardization of medical writing \cite{Park:2024,Ali:2023,Patel:2023,Cascella:2023}. They can help identify grammar errors and inconsistencies in extracted data (e.g., lab values), thereby potentially reducing documentation errors \cite{Ali:2023}. These applications may alleviate administrative burdens on healthcare professionals, allowing more time for direct patient care \cite{Lee:2023}. Nonetheless, the performance of LLMs is limited by variability in accuracy depending on case complexity, the risk of generating incorrect or fabricated content (hallucinations), and susceptibility to user framing, emphasizing the need for careful prompt design and human oversight \cite{Puthenpura:2023}.

\begin{footnotesize}
\begin{longtable}{@{}p{3.5cm}p{4cm}p{5cm}p{3cm}@{}}
\toprule
\textbf{Datasets} & \textbf{NLP Task(s)} & \textbf{Evaluation Metrics} & \textbf{Reference} \\
\midrule

Expert-annotated opioid-related posts on Reddit & Detecting addiction & Accuracy, macro-F1 & \citet{Yang2023IdentifyingSO}\\
Motivational Interviewing (MI) dataset & Generating therapeutic dialogues & ROUGE, embedding-based metrics (greedy matching, embedding average, and vector extrema), ratio of distinct n-grams, human annotator evaluation & \citet{shen-etal-2020-counseling} \\
\midrule
Empathetic Dialogues Dataset, Reddit Mental Health Dataset, DailyDialog Dataset & Generating empathetic dialogues for mental health support & BERTSCORE, accuracy, precision, recall, F-1 & \citet{Grandi2024TheES}\\
\midrule
MI-TAGS & Publicly available mental health dataset & Accuracy, macro F-1, ROC AUC & \cite{cohen-etal-2024-motivational} \\
\midrule
Reddit Self-reported Depression Diagno-
sis dataset & Interpretable NLP models in mental health & Precision, recall, F-1 & \cite{song-etal-2018-feature}\\
\midrule
Depression dataset, Non-depression dataset, Depression-candidate dataset & Interpretable NLP models in mental health & Precision, recall, F1, accuracy & \cite{zogan2021explainabledepressiondetectionmultimodalities}\\ 
\midrule

CAMS & Causal Analysis of Mental health issue & Accuracy & \cite{Garg2022CAMSAA}\\
\midrule

Dataset of principles & Simulating patient personas&  Consistency with Context, speech style, Principle Adherence & \citet{louie-etal-2024-roleplay} \\
\midrule

Publicly available depression-related conversations (RED, HOPE, ESC, AnnoMI-Full), expert-annotated preferences & Simulating patient personas & Expert evaluation on contrast with AI-like responses, linguistic authenticity, cognitive pattern authenticity, subtle emotional expression, profile adherence and personalization. Automatic evaluation on symptom severity, cognitive distortion, and overall depression Severity & \citet{liu2025eeyorerealisticdepressionsimulation}\\
\midrule

MIDAS & Publicly available dataset in mental health counseling & Expert evaluation, reflection to question ratio, accuracy, F-1 & \citet{midas} \\
\midrule

Reddit Mental Health Dataset & Publicly available mental health dataset on social media & Recall, precision, F-1 & \citet{Wu2022OntologyDrivenSF} \\
\midrule

Longitudinal Patient Health Questionnaire & Tracking user mood or mental health crises &  Spearman’s rank-order correlation, mean squared error & \citet{Gong2019MachineLD}\\
\midrule

FeedbackESConv & Providing feedback to counselors & Automatically-computed quality scores, domain experts & \cite{chaszczewicz-etal-2024-multi}\\
\midrule

PAIR, AnnoMI & Providing feedback to counselors & Edit effect (reflection score), content preservation, perplexity, coherence, specificity & \cite{min-etal-2023-verve}\\
\midrule

Anonymized counseling conversations from a NGO & Providing feedback to counselors & Adaptability, dealing with ambiguity, creativity, making progress, change in perspective & \cite{althoff-etal-2016-large} \\
\midrule

Annotated clinical notes & Predicting and understanding mental health outcomes & Accuracy & \cite{Panaite2022TheVO}\\
\midrule
Thought Records Dataset, Mental Health America & Responding to users' negative thoughts & Automatic (BLEU, ROUGE-1, ROUGE-L, BertScore); Human (Relatability, Helpfulness)\\ 

\midrule

Insomnia data set consisting from Twitter & Text classification, Correlation analysis betwen language use and insomnia, Topic modeling & True-positive rate, False-positive rate, AUC & \cite{Sakib:2021} \\

\midrule

Twitter corpus (LPHEADA) labeled for relevance to physical activity, sedentary behavior, and sleep & Text classification, Semantic consistency evaluation, location inference & Precision, Recall, F1 Score, AUC-ROC,  Average precision & \cite{Shakeri:2022} \\

\midrule

Various online recipe databases, both structured and unstructured: recipe websites, historical recipe archives, nutritional databases, sustainability data & NER, Information extraction, semantic linking, recommender system & Qualitative analysis & \cite{Erp:2021} \\

\midrule

Food Label Information and Price (FLIP) Database & Text classification, Regression, t-SNE visualization & Accuracy, Precision, Recall, F1-score, MSE & \cite{Hu:2023} \\

\midrule

Electronic health records, social media platforms (Twitter, Reddit, Facebook, YouTube), scientific literature, news and web sources & Information Extraction, Health Behavior Analysis, Early outbreak detection, Misinformation detection, Question Answering & Accuracy, Precision, Recall, F1-score (for classification), AUC-ROC, MSE (for regression) & \cite{Al-Garadi:2022} \\

\midrule 

Posts manually labeled with six annotation categories from Reddit & NER, WSD, Sequence labeling, Social media text analysis & Precision, Recall, F1-score & \cite{Hu:2021} \\

Facebook posts from antitobacco campaigns & Sentiment analysis, Topic modeling, Text classification & Odds ratios, Agreement rate & \cite{Lin:2023} \\

\midrule

Hypothetical clinical scenarios related to skin cancer & Text generation, Readability assessment & Readability score, Likert ratings & \cite{Ali:2023} \\

\midrule

Simulation of real-world clinical and research use cases & Clinical note generation, Detection of potential misuse, Language style adaptation & Qualitative analysis & \cite{Cascella:2023} \\

\midrule

Real-world pathology reports & Information extraction, Hallucination detection & Accuracy & \cite{Park:2024} \\

\midrule

12 clinical notes: 4 synthetic and 8 real & Text simplification, Definition extraction, FAQ generation, Information extraction, Prompt engineering & Quantitative evaluation of a survey, Readability score, Qualitative interviews & \cite{Mannhardt:2024} \\

\midrule

ELMTEX Dataset (clinical summaries) & Information extraction, Entity normalization & ROGUE, BERTScore, Precision, Recall, F1 & \cite{Guluzade:2025} \\

\\
\bottomrule
\caption{Overview of datasets, NLP tasks, evaluation metrics, and references from healthcare-related studies.}
\end{longtable}
\end{footnotesize}

\subsection{Education}
\paragraph{\textbf{NLP for Education Tools \& Systems.}} 
\par The integration of NLP systems into educational settings has garnered significant attention, particularly with the widespread adoption of LLMs by students. These NLP-based educational applications, such as intelligent tutoring systems, offer the potential to deliver personalized, high-quality education to underserved regions and populations. Specifically, these systems aim to enhance learning experiences by providing timely and personalized support to both teachers and students, including the following tasks: personalized and/or curriculum-aligned question generation \cite{kargupta-etal-2024-instruct,lucy-etal-2024-mathfish}, scaffolded dialogue tutoring\cite{kazemitabaar2024codeaid}, adaptive knowledge tracing \cite{kargupta-etal-2024-instruct}, automated feedback \cite{jurenka2024towards}, teacher coaching \cite{wang-demszky-2023-chatgpt}, and student simulation for testing classroom policies/activities \cite{zhang-etal-2025-simulating}.

\paragraph{\textbf{Methodologies.}} Intelligent tutoring systems primarily focused on (1) modeling teacher-student and student-student interactions using transcripts, (2) devising knowledge state spaces for specific domains/problems to trace student knowledge throughout an interaction, and (3) generating single-turn responses to students. With the emergence of LLMs, recent methodologies have expanded upon these tasks to include:

\begin{enumerate}
    \item \textbf{Multi-Turn Socratic Dialogue and Planning.} Recent methodologies leverage large language models (LLMs) to engage students in multi-turn Socratic dialogues, promoting critical thinking and problem-solving without directly providing answers. For instance, \textit{TreeInstruct} employs a state space-based planning algorithm to dynamically construct question trees based on student responses, effectively guiding learners through multi-turn code debugging tasks \cite{kargupta-etal-2024-instruct}. Similarly, the \textit{Socratic Questioning of Novice Debuggers} dataset benchmarks LLMs' abilities to employ Socratic methods in assisting novice programmers through single-turn interactions \cite{AlHossami2023}.
    \item \textbf{Expert Decision Modeling.} To emulate expert tutoring behaviors, some works model the decision-making processes of experienced educators. \textit{Bridging the Novice-Expert Gap} utilizes cognitive task analysis to capture experts' identification of student errors, remediation strategies, and instructional intentions, informing LLM responses in math tutoring scenarios \cite{wang-etal-2024-bridging}.

    \item \textbf{Curriculum-Aligned Evaluation.} Evaluating LLMs' mathematical reasoning has shifted towards alignment with educational curricula. \textit{MathFish} assesses whether models can identify and apply specific math skills and concepts as outlined in standardized curricula, using publisher-labeled data from open educational resources \cite{lucy-etal-2024-mathfish}.

    \item \textbf{Open-Ended Pedagogical Benchmarking.} To assess LLMs' instructional capabilities beyond problem-solving, \textit{MathTutorBench} introduces a benchmark evaluating open-ended pedagogical skills. It measures models' abilities across various educational tasks, emphasizing the quality of instructional interactions \cite{Macina2025}.
    
    \item \textbf{Simulated Student Interactions.} Datasets like \textit{MathDial} are created by pairing human teachers with LLMs simulating student behavior, generating rich pedagogical dialogues. This approach aids in training and evaluating models on realistic tutoring scenarios \cite{macina2023mathdial}.

    \item \textbf{Classroom Discourse Analysis.} Large-scale datasets such as the \textit{NCTE Transcripts} provide insights into teacher-student interactions. These transcripts, annotated for dialogic discourse moves, help in analyzing effective instructional practices and inform the development of NLP tools for education \cite{demszky2023ncte}.

    \item \textbf{Educational Conversation Toolkits.} Open-source frameworks like \textit{Edu-ConvoKit} facilitate the analysis of educational conversations by offering tools for preprocessing, annotation, and analysis tailored to educational research needs \cite{wang-demszky-2024-edu}.
    
\end{enumerate}

\par We have included an overview of the various dataset resources and their corresponding tasks and evaluation metrics in Table \ref{tab:edu_datasets}. These papers have been collected based on recent prominence and relevance to different challenges and opportunities present within the NLP Education space.

\onecolumn
\begin{footnotesize}
\begin{longtable}{@{}p{3.5cm}p{4cm}p{5cm}p{3cm}@{}}
\toprule
\textbf{Datasets} & \textbf{NLP Task(s)} & \textbf{Evaluation Metrics} & \textbf{Reference} \\
\midrule

MathDial & Text-to-Text Generation (Tutoring Response Generation) & sBLEU, BERTScore, KF1, Uptake, Success@k, Telling@k, Human Evaluation (Coherence, Correctness, Equitable tutoring) & \cite{macina2023mathdial} \\
\midrule
MULTI-DEBUG & Text-to-Text Generation, Text Classification (Socratic Question Generation, Multi-Turn Planning) & Relevance, Indirectness, Logical Flow, Overall Success Rate, Average \# of Turns & \cite{kargupta-etal-2024-instruct} \\
\midrule
Bridge & Text Classification, Text-to-Text Generation (Remediation of Math Mistakes, Decision-Making Modeling) & Human Evaluation (usefulness, care, human-soundingness, preference), Log Odds Ratio & \cite{wang-etal-2024-bridging} \\
\midrule
MathFish & Text Classification, Topic Modelling, Text-to-Text Generation (Math Reasoning Evaluation, Curriculum Alignment) & Weak Accuracy, Exact Accuracy & \cite{lucy-etal-2024-mathfish} \\
\midrule
MathTutorBench & Text-to-Text Generation (Evaluation of Pedagogical Capabilities in LLM Tutors) & Accuracy, BLEU, F1, Win Rate (Pedagogical skill metrics) & \cite{Macina2025} \\
\midrule
 National Center for Teacher Effectiveness (NCTE) & Text Classification (Evaluation of Classifying Educational Discourse Features) & Accuracy, Precision, Recall, F1 & \cite{demszky2023ncte} \\

\bottomrule
\caption{Overview of datasets, NLP tasks, evaluation metrics, and references from NLP for Education studies.}
\label{tab:edu_datasets}
\end{longtable}
\end{footnotesize}

\subsubsection{AI Literacy}
To identify relevant literature on AI literacy, we used Semantic Scholar and Google Scholar search terms such as ``ai literacy'' and (``ai literacy'' + ``social impact'' + ``nlp''). We found most of the papers selected for this topic to be on classroom-based studies and measurements and metrics for AI literacy, along with some interdisciplinary papers connecting AI literacy to other disciplines, for example, psychology.

\subsection{Peace Building}
We have organized highlighted papers found in our review of papers at the intersection of peace building and NLP in Table \ref{tab:peacebuilding}.

To identify the most relevant literature on human rights violation detection using NLP and on conflict prediction, we employed three search strategies: querying the ACL Anthology, conducting searches on Google Scholar, and utilizing the Consensus research discovery platform. The keywords used included: “human rights,” “human rights violations detection,” “armed conflict prediction,” and “conflict forecasting.”

The selected works on physical safety were identified through a keyword search in ACL and Google Scholar for papers on ``physical safety'', ``domestic violence'', ``gun violence'' and ``firearm injury''. The Google Scholar search also included the keyword ``nlp''.  We did not include papers that were more focused on the mental health implications of physical safety (e.g. suicide via firearms), or on larger organizational peace-building efforts (e.g. terrorism and police brutality).

We find most of these tasks in this topic to be focused on document classification of rare events. As a result, primary evaluation metrics are precision, recall, and F1. There are a few papers that apply unsupervised tasks like topic modeling and generation that use coherence and similarity, respectively, as their primary metrics.

\begin{table*}[ht]
\centering
\footnotesize
\begin{tabular}{@{}p{4cm}p{3.5cm}p{5cm}p{3cm}@{}}
\toprule
\textbf{Datasets} & \textbf{NLP Task(s)} & \textbf{Evaluation Metrics} & \textbf{Reference} \\
\midrule

Crisis Text Line Database & Document Classification (detect firearm injury or violence) &  Precision, Recall, Accuracy & \cite{Chew2023-ma}\\
National Violent Death Report System &  Topic Modeling (characterize trends in violent deaths) & Coherence, Topic diversity, Coverage & \cite{arseniev2022-violent}\\
National Electronic Injury Surveillance System Series  &  Document Classification (classify location of nonfatal gunshot injuries)  & Accuracy, Precision, Recall, AUC & \cite{parker2020-gunshot}\\
SafeText (Reddit) & Generation (generating advice) & Similarity, Confidence, Perplexity, Accuracy & \cite{levy-etal-2022-safetext} \\
Surveillance Cameras & Speech to Text, Document Classification (Detect violence)  & Precision, Recall, Accuracy, Loss & \cite{kumari2023-violence} \\
Twitter & Document Classification (detect violence related tweets) & Precision, Recall, Accuracy & \cite{ALSaif2019} \\
Police Reports & Document Classification (Classify abuse type and victim injuries) & Precision, Recall, F1 & \cite{Karystianis2019-hv} \\
Twitter & Document Classification (detect intimate partner violence) & Accuracy, F1 & \cite{ALGARADI2022-ipv} \\
Electronic Health Records & Document Classification (classify firearm injury intent) & Precision, Recall, F1 \cite{macphaul2023-firearm}\\
Gun Violence Database (deprecated compilation of news articles) & Entity Extraction (identify event details like participants, roles, location, time)   & Precision, Recall & \cite{pavlick-etal-2016-gun}\\
Twitter & POS tagging, Machine translation, Sentiment Analysis, Document Classification (detect aggression and loss) & Precision, Recall, F1 \cite{blevins-etal-2016-automatically}\\
Twitter & Document Classification (detect aggression, loss, and substance use) & Precision, Recall, F1, Average Precision & \cite{Blandfort2019-gang} \\
Twitter & Document Classification (detect aggression and loss) & Precision, Recall, F1 \cite{chang-etal-2018-detecting}\\
Electronic Health Records & Document Classification (If a patient will become violent) & F1, Confusion Matrices  & \cite{BORGER2022-psychiatric}\\
Electronic Health Record & Document Classification (type of violence and patient status) & Precision, Recall, F1 & \cite{Botelle2022-violence} \\
\bottomrule
\end{tabular}
\caption{Overview of Peace Building and Physical Safety related NLP studies}
\label{tab:peacebuilding}
\end{table*}

\subsection{Poverty}

We began our review of papers focused on the study of poverty by through keyword searches such as ``poverty detection" on Google Scholar. However, the majority of these studies addressed this issue through the use of satellite imagery~\cite{tingzon2019mapping, ayush2020generating}, or audience estimates from advertising platforms~\citep{fatehkia2020relative}. From these sets of papers, we included a review of the use of AI for poverty prediction~\cite{usmanova2022utilities}. We then modified our keyword searches on Google Scholar to ``nlp income", ``nlp poverty", and ``text poverty classification" to ensure a focus on studies from the NLP domain. For papers related to our task, we would investigate studies that have cited them as well. We have highlighted a number of the most prominent papers found in our review in Table \ref{tab:poverty}. 

The current NLP literature on the topic of poverty and class is rather limited~\cite{cercas-curry-etal-2024-impoverished}. A literature review published in 2024, only found 20 NLP papers that investigate socio-economic status in any capacity~\cite{cercas-curry-etal-2024-impoverished}. Another review of artificial intelligence systems for the detection of poverty~\cite{usmanova2022utilities}, conducted in 2022, identified 22 papers, only one of which used NLP models~\cite{muneton2022classification}. Additionally, most of the datasets are not publicly available, hindering reproduction and progress in this domain.

\begin{table*}[ht]
\centering
\footnotesize
\begin{tabular}{@{}p{4cm}p{3.5cm}p{5cm}p{3cm}@{}}
\toprule
\textbf{Datasets} & \textbf{NLP Task(s)} & \textbf{Evaluation Metrics} & \textbf{Reference} \\\midrule
Twitter & Yearly Income Prediction & Pearson correlation, Mean Average Error & \cite{preoctiuc2015studying}\\ 
Twitter & Income Prediction, Temporal Orientation Classification & Accuracy, Precision, Recall, F1, MAE & \cite{hasanuzzaman2017temporal}\\
Interview Transcripts & Poor or Extremely Poor Classification & Accuracy, Specificity, Sensitivity, F1 & \cite{muneton2022classification}\\
News & ESI Prediction, Unemployment Prediction & RMSE & \cite{lampos-etal-2014-extracting}\\
\bottomrule
\end{tabular}
\caption{Overview of Poverty related NLP studies}
\label{tab:poverty}
\end{table*}

\subsection{Online Harms}

Initial research primarily focused on developing classification frameworks across different spectrum of online harms based on existing AI models. Starting from traditional ML tools~\citep{saha2018hateminersdetectinghate}, further works exploited LSTMS/RNNs~\citep{9788347}, and with the advent of \textit{Transformers} based models (like BERT~\cite{saleh2021detectionhatespeechusing}), research rapidly unfolded with the increased development of frameworks brewed on top of attention mechanism. Recent advancements of LLMs has lead to works using their generation capability~\cite{guo2024investigationlargelanguagemodels, tassava2024developmentaiantibullyingusing, pendzel2023generativeaihatespeech}; where some works also incorporate infamous LLM strategies like zero-shot~\citep{roy-etal-2023-probing} and few-shot~\cite{zahid2025evaluationhatespeechdetection} promptings, and fine-tuning~\cite{nasir2025llmsfinetuningbenchmarkingcrossdomain}. For explainability, \citep{mathew2022hatexplainbenchmarkdatasetexplainable} presented one of the first works that proposed one-hot vector representation to improve attention based models and recently, reasoning based explanation and interpretation frameworks~\citep{yang-etal-2023-hare, nirmal-etal-2024-towards} to provide more contextual information to LLMs have garnered attention. 

Apart from traditional classification based mitigation, a rapid shift towards proactive content moderation leveraging the generative capabilities of LLMs has been proposed. Numerous works, especially focused on two strategies-- counter speech generation~\cite{bonaldi-etal-2024-nlp, saha-etal-2024-zero, wang-etal-2024-intent} and text detoxification~\cite{dementieva-etal-2025-multilingual, dale-etal-2021-text} have been extensively explored. LLMs have proven to be sufficiently good at these tasks, but need further improvements for multilingual performance~\cite{dementieva2024overview}. Some recent works have further proposed the effectiveness of strong few-shot capabilities of LLMs for annotation of such complex datasets which can potentially reduce crowd-sourcing efforts~\citep{bhat-varma-2023-large,kim-etal-2024-meganno}. Further, studies on curating multimodal datasets~\cite{kiela2021hatefulmemeschallengedetecting} and understanding the strengths and limitations of multimodal LLMs have also garnered attention~\cite{rizwan2025exploringlimitszeroshot}. We have highlighted significantly important datasets and studies on online harms in Table~\ref{tab:online_harms}. These papers are shortlisted on the basis of the impact they drive thus aiding in the detection and mitigation of online harms through necessary content moderation strategies.

\begin{table*}[ht]
\centering
\footnotesize
\begin{tabular}{@{}p{4cm}p{3.5cm}p{5cm}p{3cm}@{}}
\toprule
\textbf{Datasets} & \textbf{NLP Task(s)} & \textbf{Evaluation Metrics} & \textbf{Reference} \\
\midrule
Twitter & Multilingual and Multi-Aspect Hate speech detection spanning different target communities & Micro-F1 and Macro-F1 score & \cite{ousidhoum-etal-2019-multilingual}\\
Twitter + Gab (HateXplain) & Hate, Offensive and Normal speech detection with rationales & \textbf{Classification} - Accuracy, Macro-F1 score, AUROC score; \textbf{Rationales} - IOU-F1 score, token-F1 score, AUPRC score \textit{(Plausibility)} and comprehensiveness, sufficiency \textit{(Faithfulness)} & \cite{mathew2022hatexplainbenchmarkdatasetexplainable}\\
LLM generated explanations & Explainable hate speech detection with step-by-step reasoning generated by LLMs & Accuracy and F1 score & \cite{yang-etal-2023-hare}\\
Hate-COT & Offensive speech label explanation generated by GPT-3.5 Turbo & F1 score, Persuasiveness and Soundness & \cite{nghiem-daume-iii-2024-hatecot}\\
Latent Hatered & Detection of implicit hate speech & Precision, Recall, F1 score and Accuracy & \cite{elsherief-etal-2021-latent}\\
Jigsaw toxicity datasets & Detection of different types of toxicity across multiple labels with corresponding severity score  & Overall AUROC and Bias AUROC & \cite{jigsaw-toxic-comment-classification-challenge, jigsaw-unintended-bias-in-toxicity-classification, jigsaw-multilingual-toxic-comment-classification}\\
Measuring hate speech (Comments from YouTube, Reddit and Twitter) & Rasch Measurement Theory (RMT) based continuous scoring of hate speech across multiple labels and targets & Hate speech score, difficulty of survey item and response, severity of rater & \cite{sachdeva-etal-2022-measuring}\\ \hline \\
CONAN and its variants & Generation of counter speech against hate speech through different NLP generation strategies & Semantic Similarity, Novelty, Diversity, Toxicity, Politeness, Intent Accuracy and Hate Mitigation & \cite{chung-etal-2019-conan,fanton-etal-2021-human, bonaldi-etal-2022-human, gupta-etal-2023-counterspeeches}\\
Toxic instances from Jigsaw, Reddit and Twitter & Toxicity classifier and generation of detoxified speech for toxic instances & Accuracy, Fluency, Similarity and Joint score & \cite{logacheva-etal-2022-paradetox}\\
Multilingual text detoxification dataset from multiple sources & Multilingual text detoxification with explanation & Style Transfer Accuracy (STA), Fluency (ChrF1), Content Similarity and Joint score & \cite{dementieva-etal-2025-multilingual}\\ \hline \\
Facebook + Twitter memes & Hate, harm and misogyny detection in memes using different multimodal applications of NLP & Accuracy, Macro-F1 score and AUROC score & \cite{kiela2021hatefulmemeschallengedetecting, pramanick-etal-2021-momenta-multimodal, fersini-etal-2022-semeval}\\
BitChute videos & Hateful videos classification at the intersection of NLP, vision and audio & Accuracy, Macro-F1 score, Precision and Recall & \cite{das2023hatemmmultimodaldatasethate}\\
\bottomrule
\end{tabular}
\caption{Representative datasets, NLP task(s) and evaluation metrics from research on online harm.}
\label{tab:online_harms}
\end{table*}

\subsection{Misinformation}
\textbf{Methodology.} Automated fact-checking process comprises the verification of claims - verifiable factual statements~\citep{PANCHENDRARAJAN2024100066}. Four major components of the fact-checking pipeline~\citep{DAS2023103219} are widely studied~\citep{vlachos-riedel-2014-fact, thorne-vlachos-2018-automated, 10.1007/978-3-030-58219-7_17, guo-etal-2022-survey} and include: (1) claim detection, checkworthiness, and prioritisation (based on their urgency or potential harm / impact) to identify claims from news and social media that should be processed given the limited human and automated fact-checking resources~\citep{10.1145/3412869, 10.1007/978-3-030-72240-1_75, DBLP:journals/peerj-cs/AbumansourZ23}, often treated as a classification task; (2) evidence retrieval to collect trustworthy evidence for a claim~\citep{thorne-etal-2018-fever, augenstein-etal-2019-multifc}; (3) veracity prediction based on this evidence; and, finally, (4) explanation of the outcome label for humans~\citep{10.1145/3292500.3330935, kotonya-toni-2020-explainable, atanasova-etal-2020-generating-fact, lu-li-2020-gcan}: summarizing the evidence, generating explanations and evaluating them. The existing datasets align with the fact-checking stages. They are included in Table~\ref{tab:misinformation}. Other close tasks from the automated misinformation detection field, such as stance detection, rumor detection and fake news detection based on linguistic features are also included in this Table. In addition to this, there also more domain-specific datasets for misinformation detection - for example, multiple datasets were collected on COVID-19 topic~\citep{abdul-mageed-etal-2021-mega, song2021_covid, mohr-etal-2022-covert, heinrich-etal-2024-automatic}, including a multilingual dataset for a shared task to predict fact-checking options for claims, including their verifiability and potential harm~\citep{shaar-etal-2021-findings}. 

Misinformation can be spread in various languages, but most datasets are still in English. Multilingual verification can use translation systems, but datasets in specific languages and multilingual datasets are still needed to train and evaluate monolingual and multilingual models - for example, comprehensive multilingual multitopic claim detection datasets~\citep{PANCHENDRARAJAN2024100066} (despite the recent efforts e.g. in~\citep{10.1007/978-3-030-72240-1_75, kazemi2022, pikuliak-etal-2023-multilingual}).

For our review, we firstly focused on surveys on automated misinformation detection~\citep{guo-etal-2022-survey,DAS2023103219,PANCHENDRARAJAN2024100066,khiabani2024crosstargetstancedetectionsurvey, huang2025unmaskingdigitalfalsehoodscomparative, mis2, oshikawa-etal-2020-survey, mis4}, then studied papers and approaches mentioned.

\begin{table*}[ht]
\centering
 \small
 \footnotesize
 \begin{tabular}{@{}p{4cm}p{3.5cm}p{5cm}p{3cm}@{}}
 \toprule
 \textbf{Datasets} & \textbf{NLP Task(s)} & \textbf{Evaluation Metrics} & \textbf{Reference} \\
 \midrule
 Emergent & stance classification & accuracy, per-class precision and recall & \citep{ferreira-vlachos-2016-emergent} \\
 Multi-Target Stance Dataset & stance classification & macro-averaged F1 score & \citep{sobhani-etal-2017-dataset} \\ 
 PHEME & rumor detection and verification; stance classification & macro F1 score, accuracy & \citep{kochkina-etal-2018-one} \\
 RumourEval 2019 & stance towards a rumor: classification; veracity prediction: classification & macro F1 score; macro F1 score, RMSE & \citep{gorrell-etal-2019-semeval} \\
 VAST & stance classification & macro-averaged F1 score & \citep{allaway-mckeown-2020-zero} \\
 Will-They-Won’t-They & stance classification  for rumor verification & macro F1 score, unweighted avg F1, weighted avg F1 & \citep{conforti-etal-2020-will} \\ 
 STANDER & stance classification; evidence retrieval & macro-averaged precision, recall and F1 score; precision@5 and recall@5 & \citep{conforti-etal-2020-stander} \\
 COVID-19-Stance & classification & accuracy, macro average precision, recall, F1 score & \citep{glandt-etal-2021-stance} \\
 P-Stance & stance classification & F avg, macro-average of F1 score & \citep{li-etal-2021-p} \\
 ISD & stance detection classification & micro average F1 score & \citep{10.1145/3588767} \\ 
 C-STANCE & stance classification & F1 scores for 3 classes and and F1 macro & \citep{zhao-etal-2023-c} \\
 MT-CSD & stance classification & F avg & \citep{niu-etal-2024-challenge} \\
 TSD-CT & stance classification & F1 scores for each class and macro F1 score & \citep{zhu-etal-2025-ratsd} \\ 
 LIAR & fake news: classification & accuracy & \citep{wang-2017-liar} \\
 FakeNewsAMT and Celebrity & fake news: classification & accuracy, precision, recall, and F1 score & \citep{perez-rosas-etal-2018-automatic} \\ 
 Stance-annotated Reddit dataset & rumor stance and veracity prediction: classification & accuracy, F1 score & \citep{lillie-etal-2019-joint} \\
 Twitter-based dataset & classification & accuracy, average precision, ROC, F1 micro, F1 macro scores & \citep{volkova-etal-2017-separating} \\

 Claim detection dataset & claim detection: classification & precision, recall and F1 score & \citep{10.1145/3412869} \\
 MultiFC & Claim verification: classification; evidence ranking & micro F1, macro F1 & \citep{augenstein-etal-2019-multifc} \\
 Snopes-based dataset & stance classification; evidence extraction: ranking; claim validation: classification & precision, recall and F1 macro; precision @5 and recall @5; macro presicion, recall and F1 & \citep{hanselowski-etal-2019-richly} \\
 CLIMATE-FEVER & claim verification: retrieval, ranking and classification & accuracy & \citep{diggelmann-etal-2020-climate-fever} \\
 SciFact & claim verification: retrieval and classification & precision, recall, F1 score & \citep{wadden-etal-2020-fact} \\
 PUBHEALTH & veracity prediction: classification; explanation generation & precision, recall, F1 macro, accuracy; ROUGE and coherence & \citep{kotonya-toni-2020-explainable-automated} \\
 COVID-Fact & evidence retrieval, claim verification: classification & COVID-FEVER Score (similar to FEVER score) & \citep{saakyan-etal-2021-covid} \\
 X-Fact & claim verification: classification & F1 score & \citep{gupta-srikumar-2021-x} \\
 FakeNewsNet &  claim verification: classification & precision, recall, accuracy, F1 score & \citep{DBLP:journals/corr/abs-1809-01286} \\
 FEVER & claim verification: retrieval, ranking and classification & accuracy, F1 score; FEVER score (includes evidence retrieval and claim labels) & \citep{thorne-etal-2018-fever, thorne-etal-2018-fact} \\
 FEVEROUS & claim verification: retrieval, ranking and classification & FEVEROUS score (includes evidence retrieval and claim labels) & \citep{aly-etal-2021-fact} \\

 Multilingual claim matching dataset & claim matching: retrieval and classification & MAP@k, MRR and F1 score, accuracy & \citep{kazemi2022matchingtweetsapplicablefactchecks} \\
 CLEF-2022 CheckThat! Task 2 & claim matching: ranking & MAP, reciprocal rank, Precision@k, MAP@5 & \citep{DBLP:conf/clef/NakovMASMB22} \\
 MultiClaim & claim matching: ranking & S@10 & \citep{pikuliak-etal-2023-multilingual} \\
 NLP4IF-2021 & claim detection: classification & precision, recall, F1 score & \citep{shaar-etal-2021-findings} \\
 CLEF-2022 CheckThat! Task 1 & verifiable claim detection: classification & F1 score, accuracy, weighted F1 & \citep{10.1007/978-3-030-72240-1_75} \\
 CLEF-2024 CheckThat! Task 5 & evidence retrieval; rumor classification & MAP and Recall@5; F1 macro and strict F1 macro & \citep{DBLP:conf/clef/HaouariES24} \\
 \bottomrule
 \end{tabular}
 \caption{Datasets on misinformation detection.}
 \label{tab:misinformation}
 \end{table*}

\subsection{Inequalities and Bias}

\begin{table*}[ht]
\centering
\footnotesize
\begin{tabular}{@{}p{4cm}p{3.6cm}p{5cm}p{3cm}@{}}
\toprule
\textbf{Datasets} & \textbf{NLP Task(s)} & \textbf{Key Evaluation Metrics} & \textbf{Reference} \\
\midrule
GloVe, Word2Vec embeddings on Google corpus & Intrinsic bias probing & WEAT, gender–direction cosine & \citet{bolukbasi2016man,caliskan2017semantics} \\ 
\textsc{WinoBias} & Coreference resolution & F1 / precision gap (Female vs Male) & \citet{zhao2018gender} \\ 
\textsc{WinoMT} & Machine translation (pronoun gender) & Gender accuracy, error rate & \citet{stanovsky2019evaluating} \\ 
\textsc{BiosBias} (LinkedIn biographies) & Occupation classification & F1 diff., error disparity & \citet{de2019bias} \\ 
LLM‑Agent Interaction Logs & Multi‑agent bias propagation & Bias‑Score, fairness gap & \citet{borah-mihalcea-2024-towards} \\ \midrule
BiosBias+GPT‑J & Causal weight‑editing for debiasing & Stereotype score, accuracy & \citet{cai2024locating} \\ 
TED‑Talk / IWSLTEn–It MT & Pronoun‑drop attribution maps & Pronoun‑drop rate, TER gap & \citet{attanasio2023tale} \\ 
Synthetic counterfactual pairs (Iter\ CDA) & Bias‑robust text classification & Bias amplification ratio, F1 & \citet{plyler2025iterative} \\ 
GeoWAC, Reddit, UN General Debates & Region-aware bias evaluation metric & Region‑aware WEAT effect size, mismatch\% & \citet{borah-etal-2025-towards} \\ 
CrowS‑Pairs (intersectional ext.) & Intrinsic bias evaluation & Bias Score (Black vs White) & \citet{guo2021detecting} \\ 
\bottomrule
\end{tabular}
\caption{Representative datasets and evaluation practices across gender-bias NLP tasks. These were some influencial datasets and papers in gender bias in NLP, covering the topics of bias detection and bias mitigation methods across static embeddings, dynamic embeddings, transformer-based LMs, and LLMs}
\label{tab:bias}
\end{table*}

\begin{table*}[ht]
\centering
\footnotesize
\begin{tabular}{@{}p{4cm}p{3.6cm}p{5cm}p{3cm}@{}}
\toprule
\textbf{Datasets} & \textbf{NLP Task(s)} & \textbf{Key Evaluation Metrics} & \textbf{Reference} \\
\midrule

GeoDE, GD-VCR, CVQA & Multimodal Captioning; Multi-Agent Collaboration & Alignment score; Completeness score; Cultural Info metric & \cite{bai-etal-2025-power} \\
XNLI, PAWS-X & Cross-lingual NLI; Paraphrase Detection & Per-culture accuracy gaps & \cite{hershcovich-etal-2022-challenges} \\
World Values Survey (WVS-7) & Survey-Response Prediction; Alignment Measurement & Similarity scores; Alignment gap per group & \cite{alkhamissi2024investigatingculturalalignmentlarge} \\
CultureBank TikTok, CultureBank Reddit & Cultural QA; Zero-Shot QA; Fine-Tuning & QA accuracy improvements; Agreement levels & \cite{shi-etal-2024-culturebank} \\
NormAd-Eti & Norm Classification (acceptable vs not) & Model accuracy vs human on explicit/abstract norms & \cite{rao-etal-2025-normad} \\
GD-VCR & Culturally-Aware Captioning & Human eval (cultural descriptiveness ratings) & \cite{Yun_2024} \\
C4 web crawl & Commonsense Extraction; Classification; Clustering & Crowdsourced plausibility (PLA, COM, DIS); QA priming gains & \cite{Nguyen-2023} \\
Dollar Street & Zero-Shot Image–Text Alignment & Median CLIP score by income quartile; Spearman $\rho$ & \cite{nwatu-etal-2023-bridging} \\
Universal Dependencies treebanks; SIGMORPHON; WMT news translation; XNLI cross-lingual NLI; TyDi QA/ SQuAD & Parsing; Inflection; MT; TTS; NLI; QA & Scaled performance utility; Global utility metrics & \cite{blasi2021systematicinequalitieslanguagetechnology} \\
ImageNet & Image Classification by Country & Accuracy gaps (US/EU vs developing regions) & \cite{shankar2017classificationrepresentationassessinggeodiversity} \\
WikiAnn; Universal Dependencies treebanks; XNLI; TyDI QA/ChAII & NER; POS; NLI; QA & Utility × Demand; Gini coefficient; Throughput and memory & \cite{khanuja2023evaluatingdiversityequityinclusion} \\
Google Street View images; American Community Survey data; Voting precinct results & Vehicle Detection; Attribute Classification; Demographic Regression & Correlation vs ACS; Voting prediction accuracy & \cite{Gebru_2017} \\
ImageNet; NOAA Nighttime Lights; Google Static Maps satellite imagery & Proxy Task (Night-Light Prediction); Poverty Regression & Survey correlation vs LSMS; MAE & \cite{xie2016transferlearningdeepfeatures} \\
Gold annotations from Amazon Mechanical Turk for English, Hindi, Italian, Portuguese; Wikipedia corpora & Temporal Grounding & Hour-range accuracy vs gold annotations & \cite{shwartz-2022-good} \\
FORK test set & Commonsense QA (Culinary) & Accuracy on US vs non-US probes; Statistical significance & \cite{palta-rudinger-2023-fork} \\
TV show dialogues;  Cross‑culture shows; LDC conversational corpora & Norm Extraction; Self-Verification; Grounding & AUC for grounding; human-judged best-norm selection & \cite{fung2024normsagemultilingualmulticulturalnorm} \\
Reddit corpus of 61,981 users;  Word association benchmarks & Demographic-Conditioned LM; Word Association & Perplexity; word-association accuracy & \cite{welch-etal-2020-compositional} \\
GeoMLAMA; FORK; CANDLE; DLAMA & Cultural Commonsense QA; Country Prediction; Commonsense Verification & Accuracy gaps; uniformity analysis & \cite{shen2024understandingcapabilitieslimitationslarge} \\
Concept and Application dataset & Image Transcreation; Cultural Adaptation & Human evaluation (relevance; meaning preservation) & \cite{khanuja2024imagespeaksthousandwords} \\
WorldCuisines & Multilingual VQA (Dish Prediction; Origin Prediction) & Accuracy; adversarial context drop & \cite{winata2025worldcuisinesmassivescalebenchmarkmultilingual} \\
CVQA & Multilingual Visual QA & Accuracy; answer-matching metrics; performance drop analysis & \cite{romero2024cvqaculturallydiversemultilingualvisual} \\
LAION, GeoDE, DollarStreet & Annotation Suggestion; Data Selection & Annotation cost vs quality; coverage of cultural features & \cite{ignat2024annotationsbudgetleveraginggeodata} \\
Value-relevant outputs from 8 LLMs; Reference human value distributions from surveys & Value Alignment Probing; Distribution Mapping & Correlation with survey ground truth & \cite{cahyawijaya2025highdimensionhumanvaluerepresentation} \\

\bottomrule
\end{tabular}
\caption{Representative datasets and evaluation practices across cultural bias NLP tasks.}
\label{tab:culturebias}
\end{table*}

To curate a representative set of datasets and evaluation strategies for gender bias in NLP, we selected papers that span a wide spectrum of model architectures (from static embeddings to transformer-based LMs) and task settings (e.g., coreference resolution, occupation classification, translation, multi-agent interactions). The selection emphasizes both foundational work and recent advancements that shaped current methodologies. Early studies such as those by~\cite{bolukbasi2016man} and~\cite{caliskan2017semantics} were included for their role in establishing intrinsic bias probing techniques like WEAT. We also incorporated task-specific evaluations such as pronoun-drop metrics~\cite{stanovsky2019evaluating} and fairness gaps in classification tasks~\cite{de2019bias}. Recent papers were chosen to highlight emerging directions in LLM-based analysis, including causal interventions~\cite{cai2024locating}, region-aware bias evaluations~\cite{borah-etal-2025-towards}, and multi-agent propagation frameworks~\cite{borah-mihalcea-2024-towards}. Collectively, these works offer a diverse yet focused lens into how gender bias manifests and is measured in modern NLP systems.

We present a subset of representative datasets and evaluation benchmarks for cultural bias in Table \ref{tab:culturebias}.  

To identify relevant datasets and benchmark efforts addressing the needs of people in underrepresented communities, like people with accessibility needs, we reviewed domain-specific surveys and high-impact papers (e.g., based on citation count and venue of publication). To focus on specific disability dataset, the literature was retrieved using search queries such as \textit{“NLP accessibility datasets”}, \textit{“speech recognition for dysarthria”}, \textit{“text simplification benchmark”}, and \textit{“sign language translation dataset”} in Google Scholar and ACL Anthology. When disability-specific datasets were not available, we included in the table the most widely used datasets for the corresponding NLP task. We present a representative collection of works in Table \ref{tab:under}.

\begin{table*}[ht]
\centering
\footnotesize
\begin{tabular}{@{}p{4cm}p{2.6cm}p{5cm}p{3cm}@{}}
\toprule
\textbf{Datasets} & \textbf{NLP Task} & \textbf{Evaluation Metrics} & \textbf{Reference} \\
\toprule
Newsela, WikiLarge, \\ WikiSmall, ASSET, MUSS, \\ SimplicityDA, Simple Wiki & Text Simplification & SARI, FKGL, BLEU, BERTScore, Human Ratings & \citet{al2021automated}\\

\midrule
LJSpeech, VCTK, LibriTTS, Blizzard Challenge, HiFi-TTS, M-AILABS, CSS10, AISHELL & Text-to-Speech & MOS (Mean Opinion Score), Intelligibility Score, Naturalness, Word Error Rate (WER), MCD (Mel-Cepstral Distortion) & \citet{khanam2022text,kumar2023deep}\\

\midrule
EasyCall, UASpeech, TORGO, CSLU Dysarthric, DEED, \\ L2-ARCTIC, Google Project Euphonia & Speech Recognition & WER (Word Error Rate), CER (Character Error Rate), Accuracy, Intelligibility, Real-time Factor (RTF) & \citet{alharbi2021automatic,li2022recent}\\  

\midrule
DSBI, Smart Braille Converter Corpus, Tamil-Braille Dataset & Braille Processing & Accuracy, Precision, Recall, BLEU (for translation), OCR Error Rate & \citet{ali2023artificial}\\  

\midrule
MS COCO, VizWiz, Multi30K, Flickr8k, Flickr30k, STAIR Captions, TextCaps, OpenSubtitles & Image Captioning and Subtitling & BLEU, METEOR, ROUGE, CIDEr, SPICE, Human Ratings (fluency, adequacy), Caption Accuracy & \citet{stefanini2021show,ghandi2023deep}\\  

\midrule
VizWiz VQA, TDIUC, VQA-Med, OK-VQA, GQA, TextVQA, DocVQA, OViQA, PathVQA, & Question Answering & Accuracy, VQA Score, BLEU, ANLS (Average Normalized Levenshtein Similarity), Human Ratings & \citet{gurari2018vizwiz, chen2022grounding,huh2024long} \\ 

\midrule
PHOENIX14T, RWTH-PHOENIX-Weather, CSL-Daily, RWTH-BOSTON-104, ASLG-PC12, OpenASL, RWTH-SLT, Sign2Text, How2Sign, AUTSL & Sign Language & BLEU, ROUGE, METEOR, WER, Sign Error Rate (SER), Gloss Accuracy, Human Ratings & \citet{yin2021including, tao2024sign}. \\  
\bottomrule
\end{tabular}
\caption{Representative datasets and evaluation practices across accessibility-related NLP tasks.}
\label{tab:under}
\end{table*}

\subsection{Environmental Harms}
Given the absence of a survey paper in this field, we began our review with research presented at the inaugural \href{https://aclanthology.org/events/climatenlp-2024/}{Workshop on Natural Language Processing Meets Climate Change (ClimateNLP 2024)}. Additionally, we have included follow-up studies conducted by researchers within the ClimateNLP community. A significant portion of the work in this domain focuses on classifying climate-related claims, text, or stances. Detecting misinformation has emerged as a prominent topic, often tackled via external source verification and question-answering approaches. Information extraction, whether related to quantitative features or narrative insights, also plays a key role. The table \ref{tab:environment} below highlights the key papers associated with these tasks.

\begin{table*}[ht]
\centering
\footnotesize
\begin{tabular}{@{}p{4cm}p{3.5cm}p{5cm}p{3cm}@{}}
\toprule
\textbf{Datasets} & \textbf{NLP Task(s)} & \textbf{Evaluation Metrics} & \textbf{Reference} \\

\midrule
Global Stocktake Dataset from Climate Policy Radar & Topic modelling & Cosine similarity & ~\cite{sietsma-etal-2023-progress} \\
\hline
SumIPCC: topic-annotated summaries and relative paragraphs from climate change reports & Aspect-based summarisation & Mean Reciprocal Rank (MRR), Carburacy-reweighted ROUGE score & ~\cite{ghinassi-etal-2024-efficient} \\

\hline
ClimateFever: real-world climate climate change claims with evidence sentences from Wikipedia & Text classification & Label-accuracy, F1, precision, recall & ~\cite{diggelmann-etal-2020-climate-fever} \\

\hline
TCFD-category labeled sentences from firms' annual reports, sustainability-, climate- or TCFD-reports and firms' webpage & Text classification & Accuracy & ~\cite{bingler-etal-2022-cheaptalk} \\
\hline
CORP: paragraphs from common news, research articles and climate reporting of companies & Language modelling, text classification, sentiment analysis, fact-checking & Average cross-entropy loss, average validation loss, weighted F1 score & ~\cite{leippold-etal-2022-climatebert} \\
\hline
Reduction target claims collected by Net Zero Tracker & Text classification & Accuracy, precision, recall, F1 score & ~\cite{schimanski-etal-2023-climatebert-netzero} \\
\hline
HLEG reports, Net Zero Stocktake reports, Corporate Climate Responsibility Monitor Reports from NewClimate Institute & Q\&A, text generation & Expert evaluation scores for quality, factual accuracy, relevance & \cite{hsu-etal-2024-evaluating} \\
\hline
Corpus created from curated sources compiled by ERASMUS.AI (Pretraining), collection of scientific reports and papers (RAG), ClimaBench (Downstream tasks): collection of climate-related datasets for classification & Language modelling, Q\&A, various downstream tasks (classification etc.)  & Cross-entropy loss, average validation loss, weighted F1-score, BLEU scores, human evaluation & ~\cite{thulke2024climategptaisynthesizinginterdisciplinary} \\
\hline
Sixth Assessment Reports (AR6) of IPCC & Q\&A, text generation & Accuracy score given by experts & ~\cite{vaghefi-etal-2023-chatclimate} \\
\hline
Text from IPCC, WMO, AbsCC (climate change abstracts), 1000S (abstracts by top 1000 climate scientists) & Text classification & Averaged micro-F1 score with different classification levels & \cite{leippold-etal-2025-factchecking} \\
\hline
SemTabNet: tables from over 10K corporate ESG reports obtained using Deep Search toolkit & Information extraction & Tree Similarity Score & ~\cite{mishra-etal-2024-statements} \\
\hline
Corporate annual and sustainability reports & Quantitative information extraction & Custom report-level metrics evaluating retrieval and accuracy of extractions & ~\cite{dimmelmeier-etal-2024-informing} \\
\hline
ClimateQA: climate claims & Q\&A & Manual inspection, ROUGE-L recall score, conditional probability, truth ratio, GPT-Match, GPT-Contradiction, AlignScore & \cite{fore-etal-2024-unlearning} \\
\hline
Climate change-related questions from Google Trends, Skeptical Science, synthetic questions from English Wikipedia & Evaluating LLMs' responses & Presentational (style, clarity, correctness, tone) and epistemological (accuracy, specificity, completeness and uncertainty) properties & ~\cite{bulian-etal-2024-assessing} \\
\hline
Corporate sustainability reports & Text summarization, Text scoring, Q\&A & Human evaluation for hallucinations, ROUGE precision score & ~\cite{ni-etal-2023-chatreport} \\
\hline
Climate-related questions (question-source-answer pairs), sustainability report dataset (report-paragraph-question pairs) & Information extraction & Recall@K, Precision@K, F1@K for different top K values & \cite{schimanski-etal-2024-climretrieve} \\

\hline 
Dataset of corporate climate policy engagement documents collected by LobbyMap &   Information extraction (query, stance and evidence page indices), classification & Strict F-score, page overlap F-score, document F-score, & \cite{NEURIPS2023_7ccaa4f9} \\
\hline
Facebook ads related to climate change (oil and gas sector), text and spend, impressions, demographic and regional distribution & Multi-label classification & Overall and sub-category specific F-score & \cite{holder-etal-2023-climate, rowlands-etal-2024-predicting} \\
\hline
News articles in English and Mandarin & Information extraction (narrative features) & Human evaluation, ROUGE-1, ROUGE-L, cosine similarity & \cite{zhou-etal-2024-large} \\

\bottomrule

\end{tabular}
\caption{Selection of influential datasets and papers in the ClimateNLP domain, ranging from topic modelling over text classification to question answering and information extraction}
\label{tab:environment}
\end{table*}

\clearpage

\section{Global Goals}
\begin{figure*}[h!]
    \centering
    \includegraphics[width=0.9\linewidth]{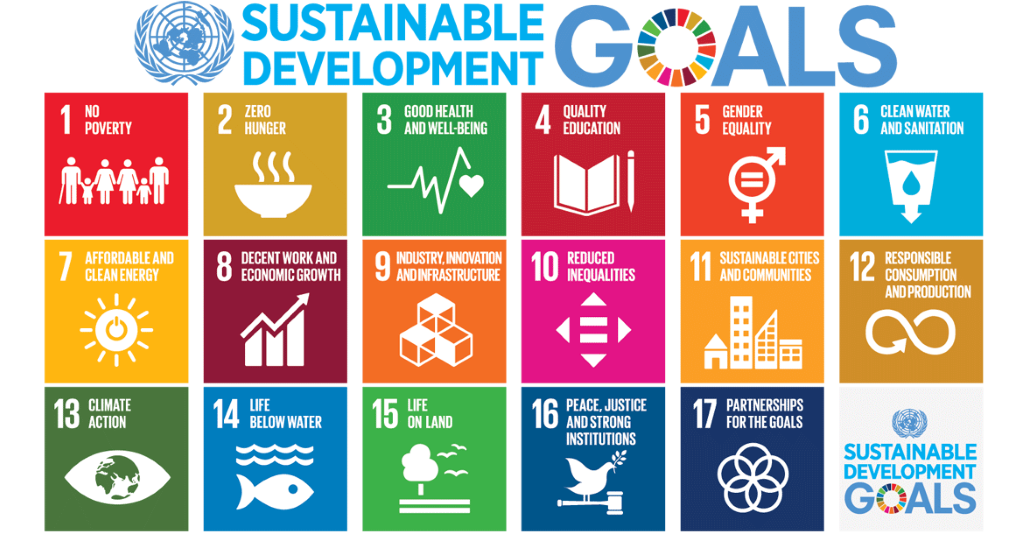}
    \caption{Overview of the SDG goals. Source: \href{https://sdgs.un.org/goals}{https://sdgs.un.org/goals}}
    \label{fig:sdgs}
\end{figure*}

We include an overview of the Sustainable Development Goals (SDGs) in Figure \ref{fig:sdgs}, which apply to many of the NLP applications discussed in this paper.

\section{Global Risks}

\begin{figure*}[h!]
    \centering
    \resizebox{0.95\textwidth}{!}{\includegraphics{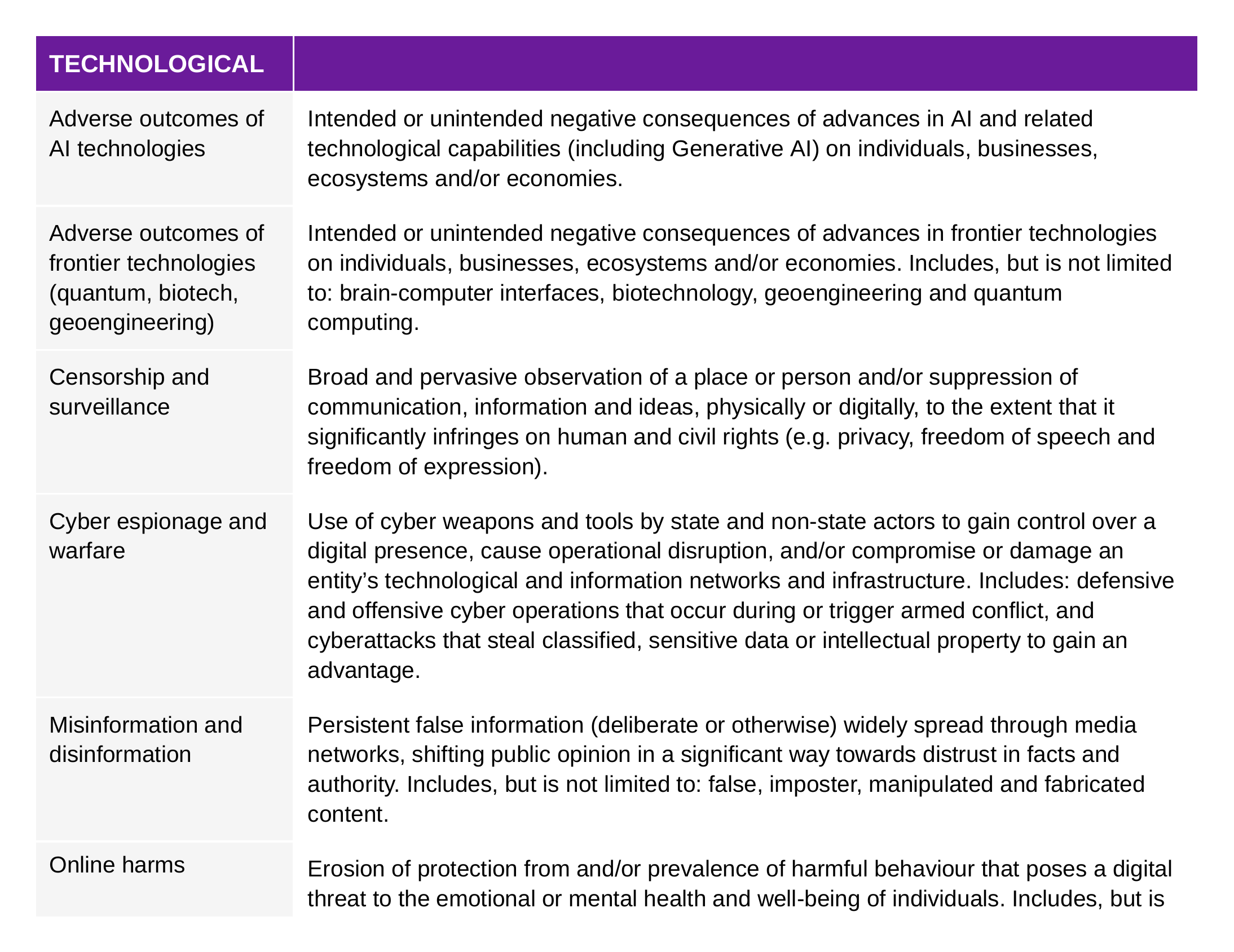}}
    \caption{Domain-specific global risks according to the \textit{Global Risks Report 2025}. Each table illustrates key risks in societal, technological, geopolitical, environmental, and economic domains respectively.}
    \label{fig:global-risks-composite}
\end{figure*}

\begin{figure*}
    \centering
    \includegraphics[width=0.99\linewidth]{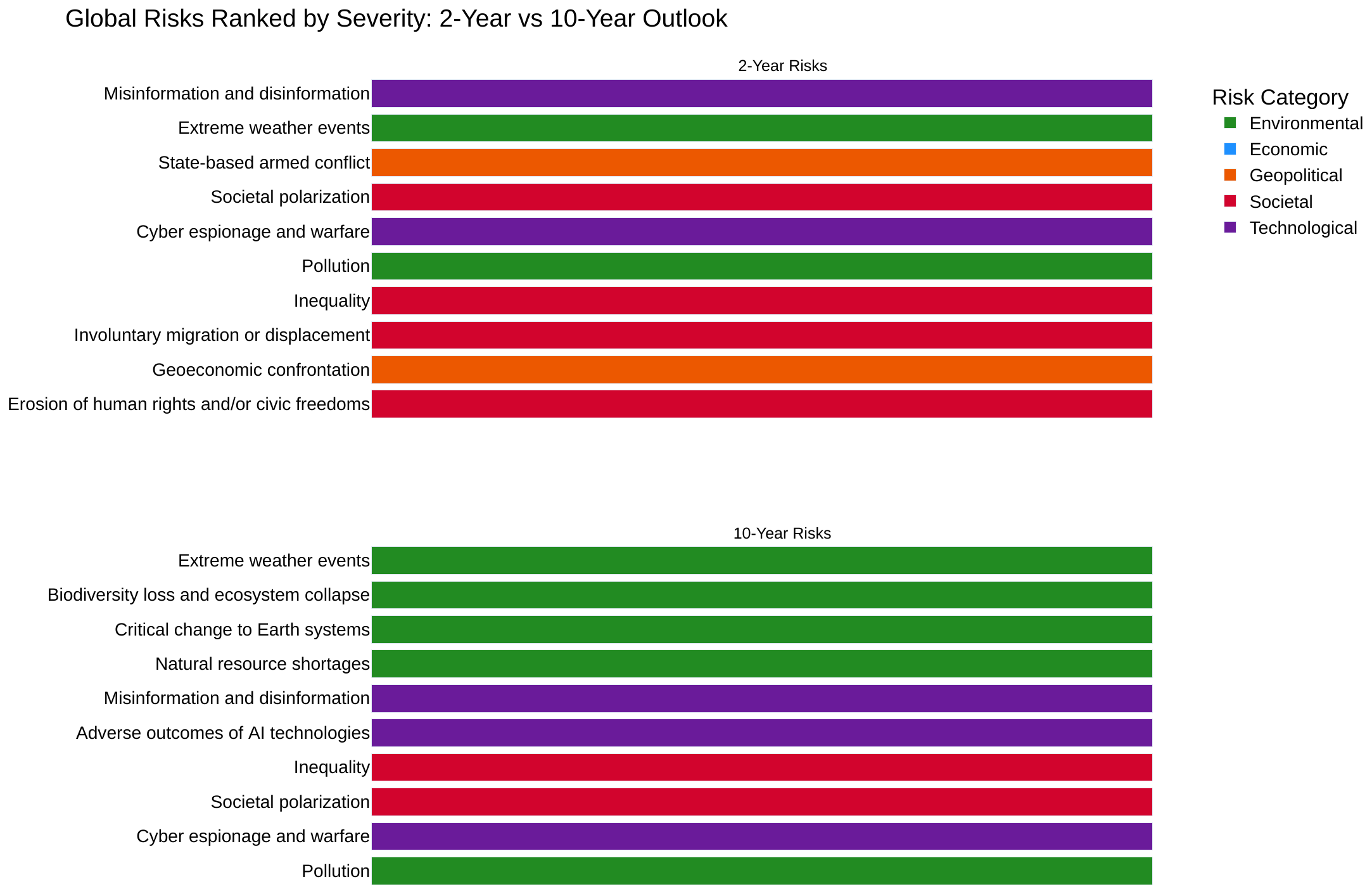}
    \caption{Global risks ranked by severity over the short and long term. Reproduction of Figure FIGURE C from: \href{https://reports.weforum.org/docs/WEF_Global_Risks_Report_2025.pdf}{the Global Risks Report 2025}}.
    \label{fig:risks_predictions}
\end{figure*}

We present the key global risks categorized by domain, as outlined in the \textit{Global Risks Report 2025} by the World Economic Forum.\footnote{\href{https://www.weforum.org/publications/global-risks-report-2025}{https://www.weforum.org/publications/global-risks-report-2025}} Each domain-specific table in Figure \ref{fig:global-risks-composite} highlights major threats along with their definitions. We also provide the global risks ranked by severity over the short and long term in Figure \ref{fig:risks_predictions}.

\begin{figure*}[h!]
    \centering
    \resizebox{0.95\textwidth}{!}{\includegraphics{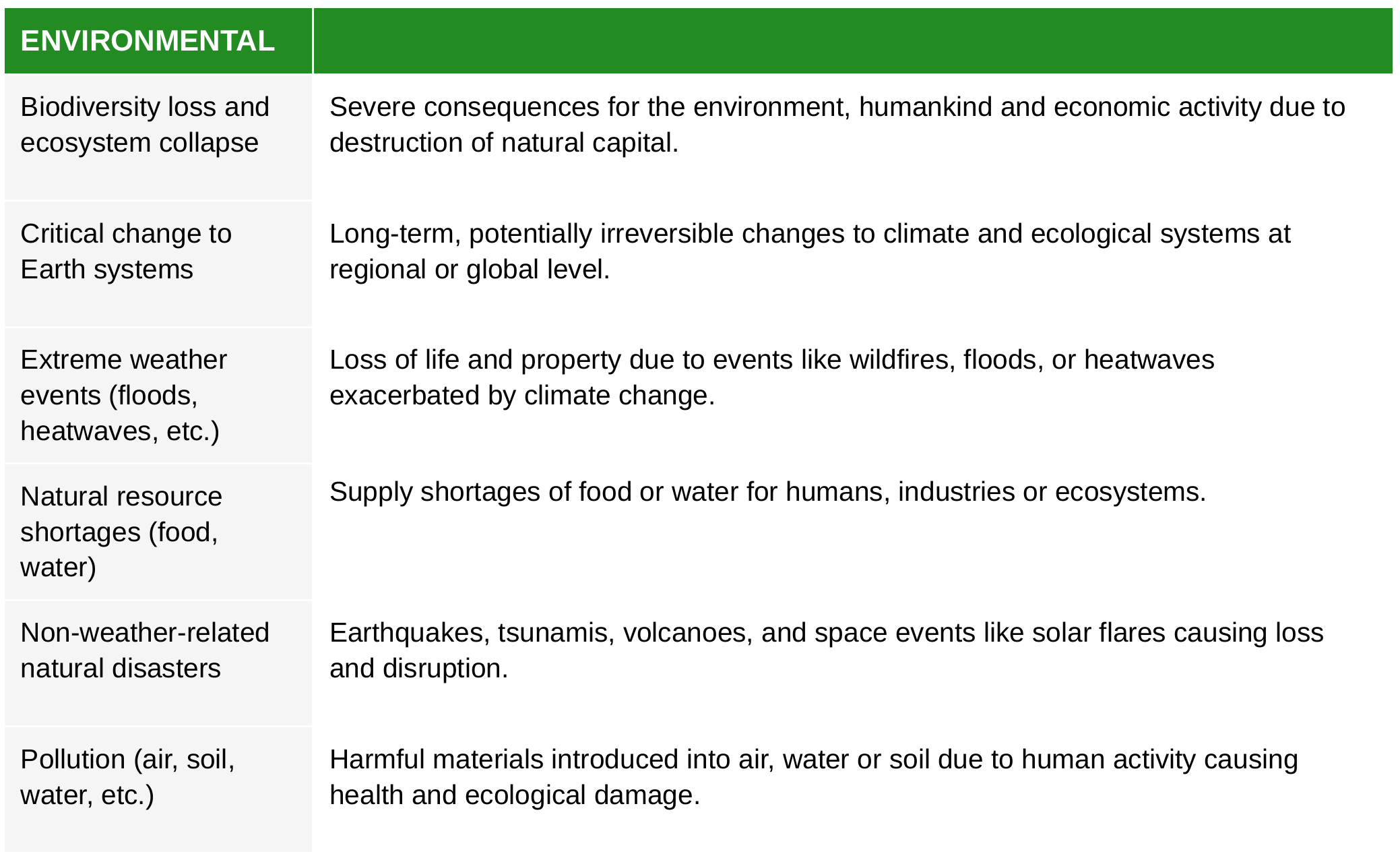}}\\[1em]
    \resizebox{0.95\textwidth}{!}{\includegraphics{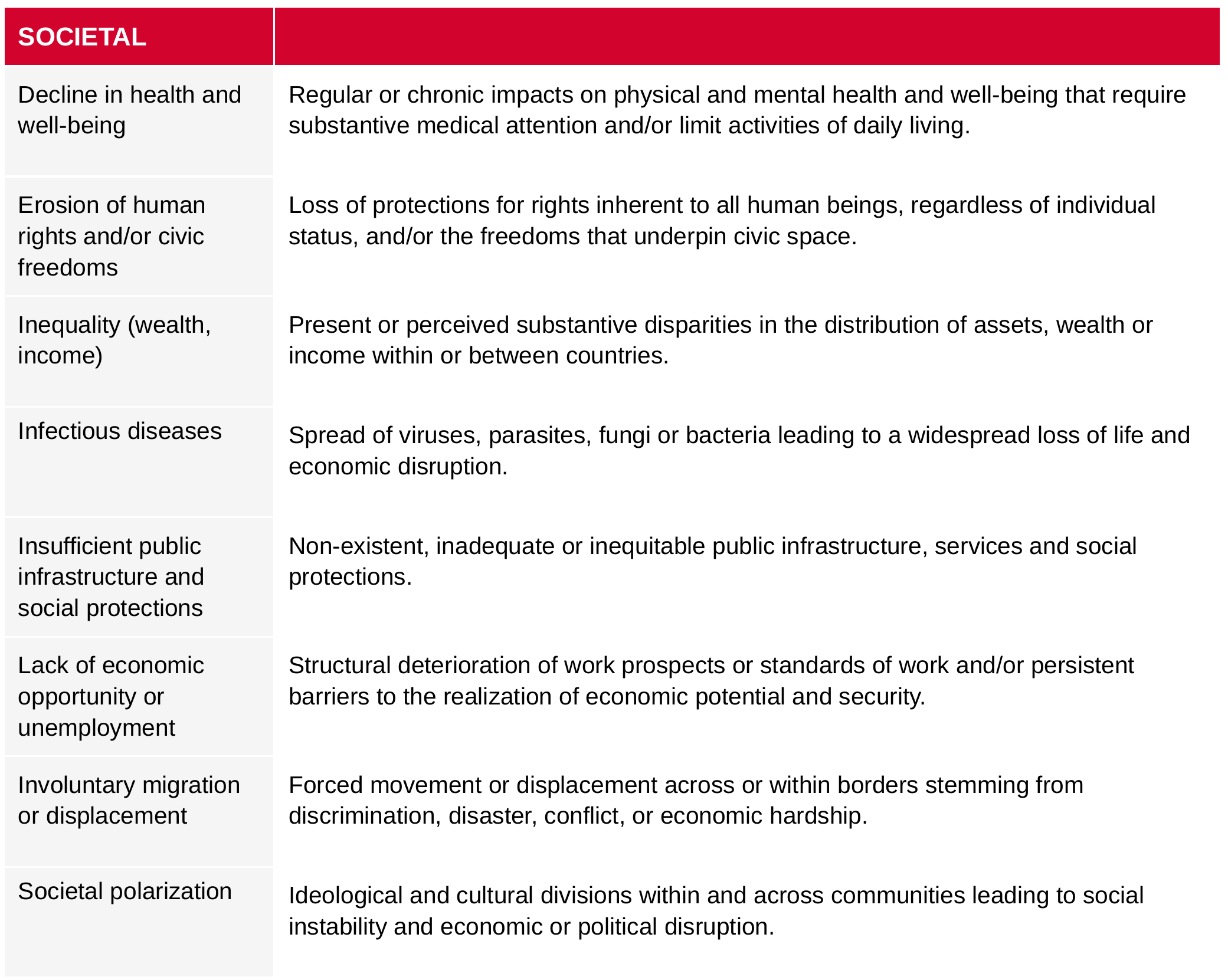}}\\
\end{figure*}

\begin{figure*}[h!]
    \centering
    \resizebox{0.95\textwidth}{!}{\includegraphics{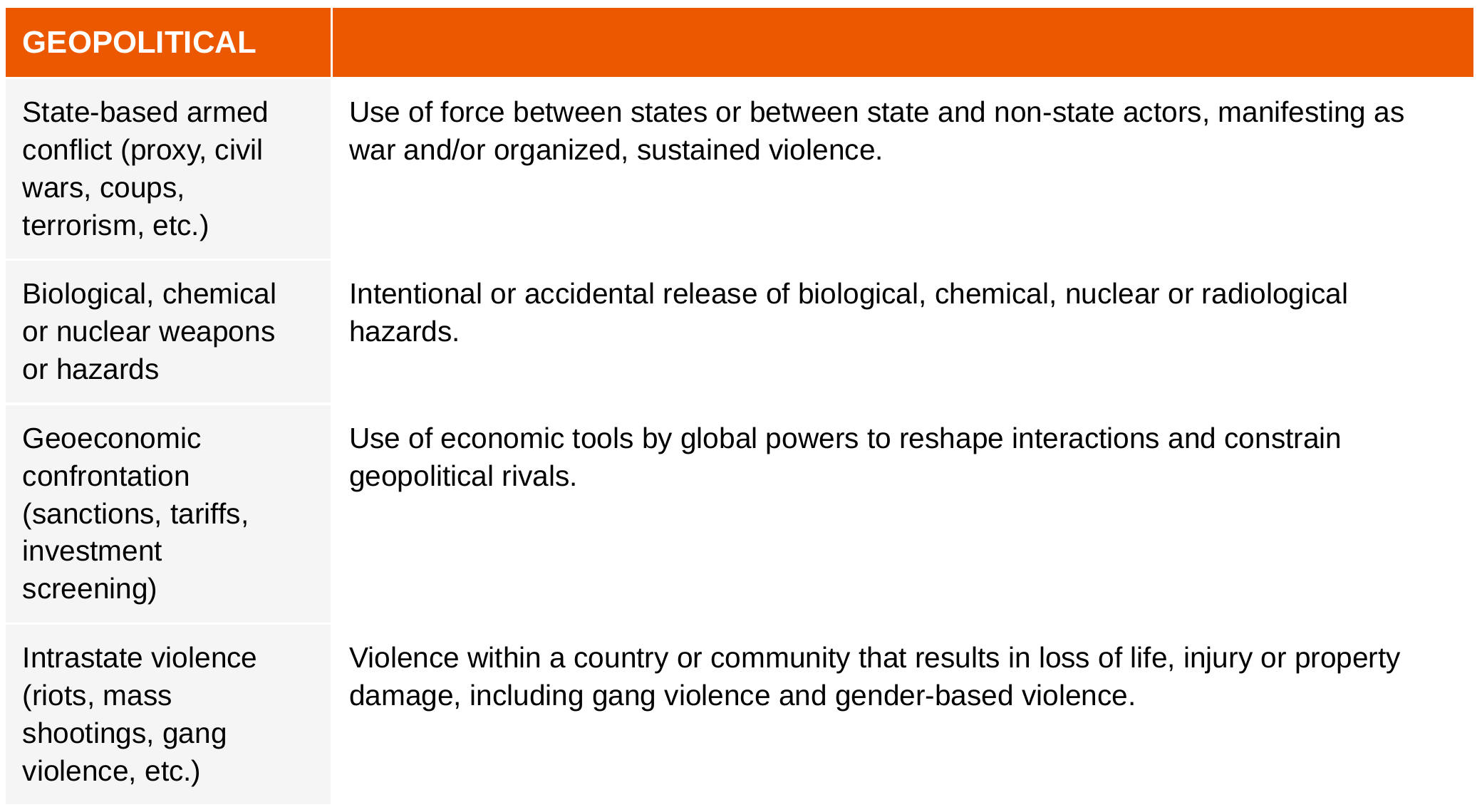}}\\[1em]

    \resizebox{0.95\textwidth}{!}{\includegraphics{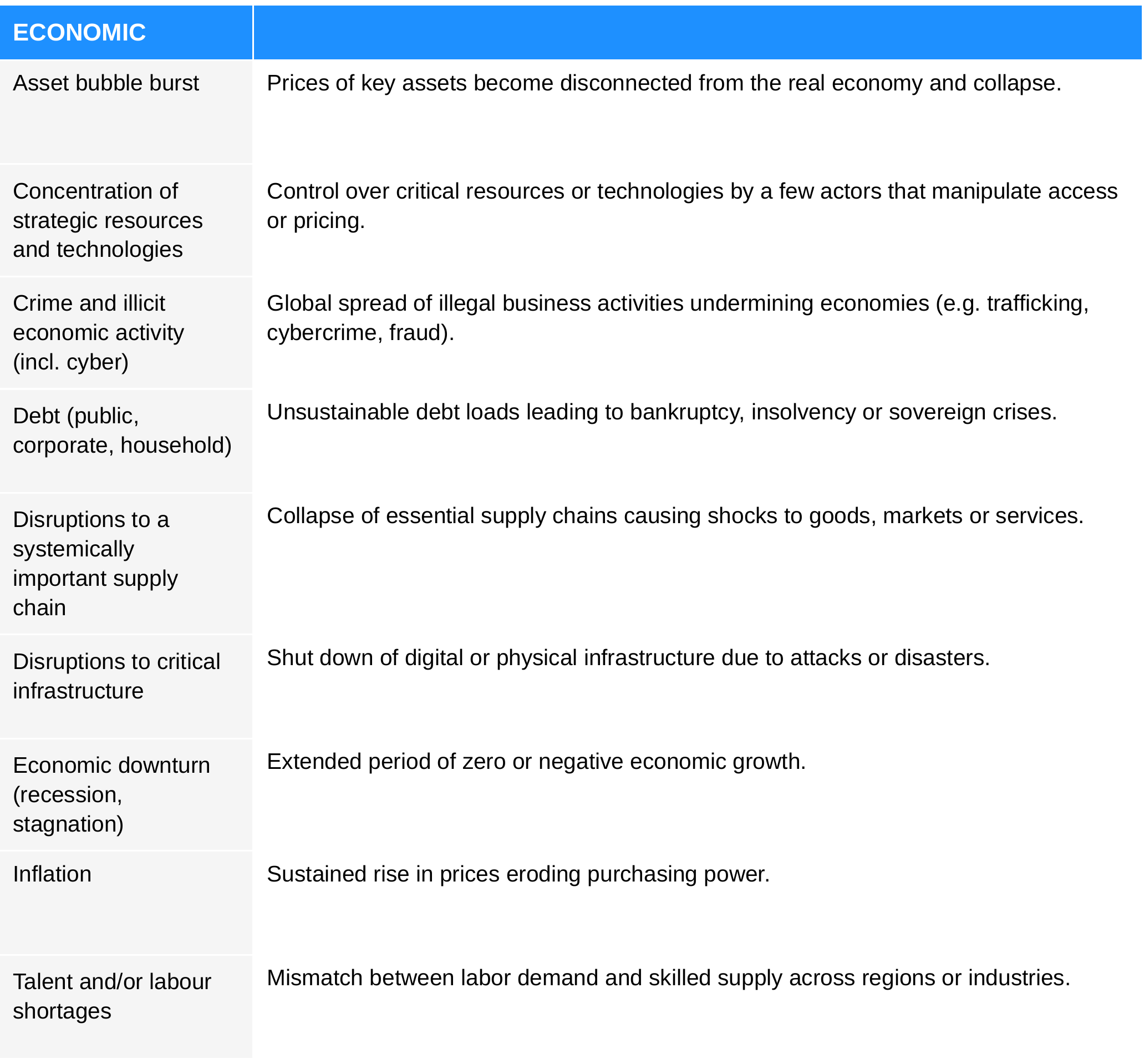}}
    \end{figure*}

\clearpage
\newpage

\section{Guidelines for authors}
This work involved contributions from many collaborators with diverse backgrounds, who surveyed various topics and guided the writing process. 
After extensive discussions in the project’s early phase, we developed the author guidelines shown in Figure \ref{tab:guidelines}. 
We share these here for transparency reasons and to assist other researchers undertaking similar multidisciplinary efforts.

\begin{figure*}[h!]
\begin{tcolorbox}[myguidelinesbox]

\begin{footnotesize}
Your task is to write a section (or subsection) based on a topic of your expertise. To ensure consistency across all sections, we will first collaboratively collect key papers, identify the needs they address, and outline the NLP methodologies they apply. This information, along with our notes, will be organised in a shared spreadsheet. Once the research is mapped out, we will proceed to draft the sections, following a consistent structure that highlights both current opportunities and existing challenges in the field.

Please work on your chosen topic tab and fill in the shared 
[link]{spreadsheet}. Below are the main suggestions to help guide your input:

\begin{itemize}
    \item \textbf{Spreadsheet Columns:}
    \begin{itemize}
        \item \textit{Main Field Papers} – Select key papers in the area. We recommend starting from potential existing surveys in the field to identify key papers. If no surveys exist, we recommend using a keyword search and manual filtering of the most influential papers using the number of citations if needed.
        \item \textit{Social Needs Covered} – Societal challenges addressed. We need this information for Section 2. So, please update the column in your tab accordingly, even if you don't use it in your section.
        \item \textit{Popular Datasets} – Commonly used datasets
        \item \textit{NLP Task(s)} – How tasks are defined (generation, classification, etc.)
        \item \textit{NLP Methodology (Existing)} – Methods used in literature
        \item \textit{Evaluation} – Evaluation setup and metrics
        \item \textit{Limitations} – Limitations of current work
        \item \textit{Challenges / Open Questions} – Remaining gaps or issues
        \item \textit{Expected NLP Impact / Suggestions} – Potential contributions and ideas
        \item \textit{NLP Methodology Potential} – Methodological insights or improvements
    \end{itemize}

    \item \textbf{Suggested structure for your section:}
    \begin{itemize}
        \item \textit{Methodology:} – Dataset, approach, evaluation
        \item \textit{Limitations:} – Challenges, Critical analysis, and Future Work
        \item \textit{Opportunities:} – Suggestions, Open Questions,  Impact, Broader relevance and Impact in NLP
    \end{itemize}

        \item \textbf{Suggested length for your topic-section:}
    \begin{itemize}
        \item \textit{We aim for three main discussions per topic: methodology, limitations, and opportunities.}
        \item \textit{Please use a table for the methodology section that will be added in the appendix.}
        \item \textit{We can aim for a maximum of 4 paragraphs for your chosen topic: general intro, methodology paragraph (going to the appendix), challenges, opportunities. In the main paper, you have one column each. Keep in mind that we might keep a shorter version in the final revisions.}
        
    \end{itemize}
\end{itemize}
\end{footnotesize}

\end{tcolorbox}
\caption{Guidelines for the authors of the paper. Please reach out for any clarification.}
\label{tab:guidelines}
\end{figure*}

\end{document}

%% file: latex/colourcodes.tex
\usepackage{xcolor}

\definecolor{SDG1}{RGB}{229,36,59}
\definecolor{SDG2}{RGB}{221,166,58}
\definecolor{SDG3}{RGB}{76,159,56}
\definecolor{SDG4}{RGB}{197,25,45}
\definecolor{SDG5}{RGB}{255,58,33}
\definecolor{SDG6}{RGB}{38,189,226}
\definecolor{SDG7}{RGB}{252,195,11}
\definecolor{SDG8}{RGB}{162,225,66}
\definecolor{SDG9}{RGB}{253,105,37}
\definecolor{SDG10}{RGB}{221,19,103}
\definecolor{SDG11}{RGB}{253,157,36}
\definecolor{SDG12}{RGB}{191,139,46}
\definecolor{SDG13}{RGB}{63,126,68}
\definecolor{SDG14}{RGB}{10,151,217}
\definecolor{SDG15}{RGB}{86,192,43}
\definecolor{SDG16}{RGB}{0,104,157}
\definecolor{SDG17}{RGB}{25,72,106}

\definecolor{R1}{RGB}{255,144,77}
\definecolor{R2}{RGB}{26,137,4}
\definecolor{R3}{RGB}{25,153,211}
\definecolor{R4}{RGB}{230,83,83}
\definecolor{R5}{RGB}{150,96,159}

%% file: latex/1_INTRO.tex
\section{Introduction}
\begin{flushright}
\footnotesize{
\textit{``Understanding the problem is half the solution.''} \\
\hfill --- Charles Kettering
}
\end{flushright}

\begin{figure}[t]
    \centering
    \includegraphics[width=\linewidth]{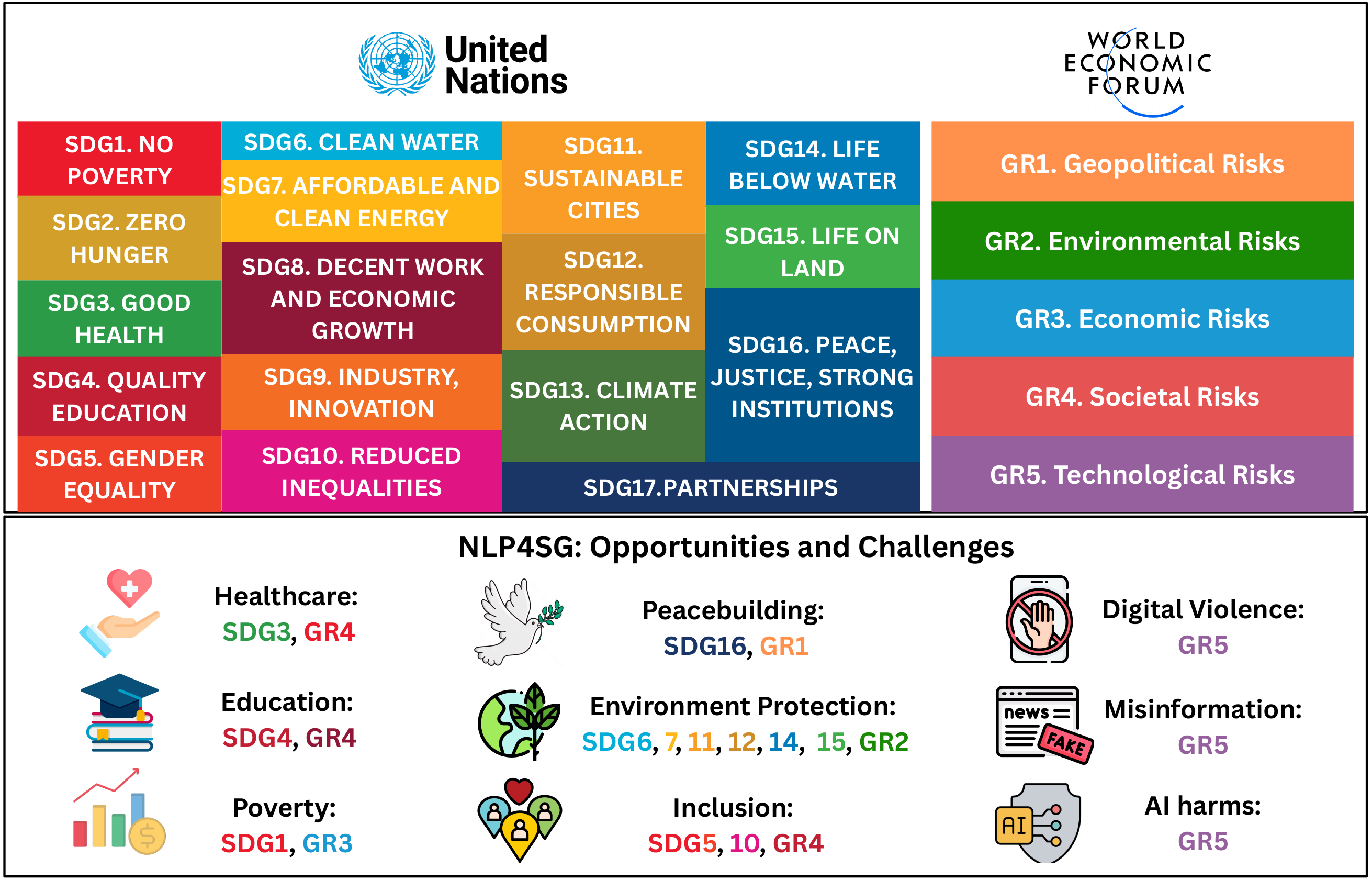}
    \caption{Mapping NLP applications for Social Good (\textbf{NLP4SG}) with global goals and risks.}
    \label{fig:overview}
\end{figure}

\noindent To fully realize the potential of NLP, it is essential to look beyond technical achievements and reframe tasks around pressing societal needs. We draw on insights from the United Nations Sustainable Development Goals\footnote{\scriptsize{\href{https://sdgs.un.org/goals}{https://sdgs.un.org/goals}}} (SDGs) and the 2025 Global Risks (GRs) Economic Report\footnote{\scriptsize{\href{https://www.weforum.org/publications/global-risks-report-2025/digest}{https://www.weforum.org/publications/global-risks-report-2025/digest}}} to provide a foundation for an interdisciplinary recontextualization of NLP, encouraging reflection on how language technologies intersect with today’s most pressing challenges.
We selected these two agendas as, from a social good perspective, UN SDGs offer a global framework for fostering peace and prosperity for people and the planet. However, while highly influential, these goals were established in 2015---prior to the rapid advancements in artificial intelligence. To contextualize them within today's technological landscape, we also draw on insights from the 2025 GR Report, which highlights both the transformative potential and the emerging global risks associated with technology and information processing. The resulting mapping of NLP application domains to SDGs and GRs is shown in Figure~\ref{fig:overview}. Our effort builds on prior research that assesses the role of NLP through positive impact \cite{hovy-spruit-2016-social, jin-etal-2021-good}, maps NLP4SG work to the SDGs \cite{adauto-etal-2023-beyond, gosselink2024aiactionacceleratingprogress},  outlines open questions in modern NLP~\cite{ignat-etal-2024-solved}, and limitations in NLP and AI pipelines \cite{mihalcea2025ai}. Thus, our research goal in this work is threefold: \textbf{RQ1}—what NLP-based solutions already support positive social impact, \textbf{RQ2}—what challenges arise in developing them, and \textbf{RQ3}—what promising directions remain overlooked?

\begin{figure}[t]
    \centering
    \includegraphics[width=\linewidth]{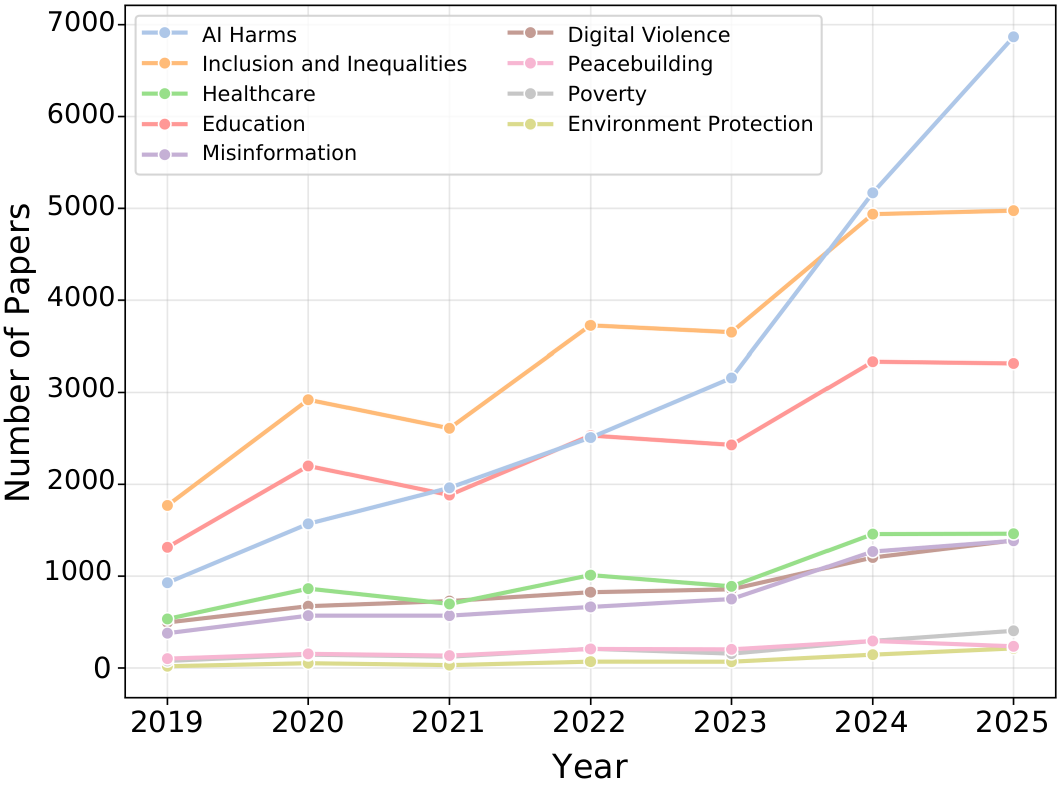}
    \caption{Number of ACL Anthology papers per domain showing publication volume and growth trajectories over time (2019–2025).}
    \label{fig:stats}
\end{figure}

We first analyze publication trends in the ACL Anthology across nine NLP research directions (Figure~\ref{fig:stats}), revealing uneven growth across domains. For each domain, we review existing work and current challenges,\footnote{Further statistics and details about survey methodology and paper selection are provided in Appendix~\ref{sec:appendix}.} and present an outlook on opportunities that reflects our perspective on how the field can advance. We conclude by synthesizing overarching research directions and community actions to encourage more proactive NLP4SG efforts.

%% file: latex/2_TOPICS.tex


\input{latex/healthcare/personal_well_being}

\input{latex/education/education}

\input{latex/economy/economic_growth}

\input{latex/peace_building_and_human_rights/peace}

\input{latex/environment/climate}

\input{latex/inequalities-inclusion/inequalities}

\input{latex/technological_risks/digital_violence}

\input{latex/misinformation/misinformation}

\input{latex/technological_risks/harms}

%% file: latex/healthcare/personal_well_being.tex
\section{\includegraphics[height=1.5em]{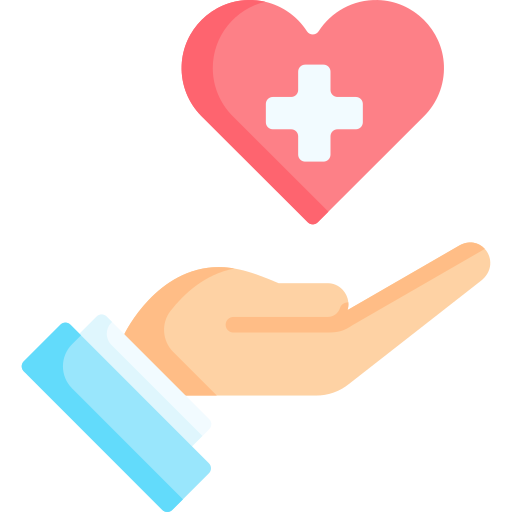} \hspace{0.1em}Healthcare}

NLP can help address challenges linked to \textcolor{SDG3}{SDG3} (Good Health and Well-being) by improving healthcare delivery and outcomes; and to \textcolor{R4}{GR4} (Societal Risks), including health decline, health workforce shortages, and infectious disease outbreaks.

\paragraph{\it Mental Health.} As healthcare access remains uneven, NLP systems, particularly LLMs, offer scalable means to reduce support gaps for those limited in time and resources \cite{tong2023conceptualizing, hua2024large}. As \textit{counselors}, they could assist in \textit{detecting} conditions like depression and addiction from clinical or social media data \cite{giuntini2020review, Yang2023IdentifyingSO}; \textit{responding} empathetically by interpreting emotion and generating therapeutic dialogues \cite{shen-etal-2020-counseling, Grandi2024TheES}; and \textit{tracking} user mood or crises over time \cite{osi2024WarEM, Gong2019MachineLD}. As \textit{clients}, NLP tools simulate diverse personas to train and evaluate counselors \cite{louie-etal-2024-roleplay, liu2025eeyorerealisticdepressionsimulation}.

\paragraph{\it Physical Health.} Prior work has focused on physical well being, using social media mining for tracking physical activity, sleep patterns \cite{Sakib:2021,Shakeri:2022}, diet habits \cite{Erp:2021,Hu:2023}, gauging public health attitudes such as mask-wearing during COVID-19 \cite{He:2021}, and detecting risky behaviors like substance use \cite{Hu:2021,Lin:2023}. In clinical settings, NLP aids in medical record analysis via classification for treatment decisions, named entity recognition for patient-trial matching, relation extraction linking symptoms and treatments, predictive modeling of treatment responses \citep{jerfy2024growing}, and information extraction to organize text reports \cite{Sheikhalishahi:2019,Landolsi:2023}.

\paragraph{Challenges:} Key challenges involve the availability of \textit{health-related data}, which is scarce, sensitive, and often biased, with limited language coverage and marginalized group representation raising ethical and privacy concerns~\citep{Ford:2019,Shakeri:2022,midas}; the \textit{evaluation framework}, which must go beyond accuracy to reflect fairness, contextual understanding, human-centered values (e.g. empathy), and ensure reproducibility; and \textit{long-term user impact}, as reliance on LLMs for sensitive tasks risks sycophancy, ELIZA effects \cite{ekbia2008artificial}, and overdependence. There is also a \textit{lack of causal frameworks} and \textit{interpretable} models, which complicates the understanding in the outcomes of NLP systems \cite{zhang2022natural}. Remote care with LLM-based tools risks diverting individuals from essential in-person treatment \cite {Khawaja2023YourRT, 10.1145/3453175}. This highlights a broader challenge: ensuring such technologies \textit{support—not replace—}human professionals, with responsible use as a guiding principle\cite{Shakeri:2022,brown2021ai}. There is growing evidence that LLMs lack the stability, contextual grounding, and ethical guarantees required for autonomous clinical or therapeutic decision-making~\citep{zhao-etal-2024-llms, Iftikhar, Cui2025-jm}, highlighting the need for informed, cautious, and supervised use of such tools.

\paragraph{Opportunities:} Future work should continue to emphasize \textit{multimodal approaches}  that jointly leverage text, speech prosody, facial expressions, physiological signals, and other data sources like sensors, wearables, or images for richer context and personalization \cite{Puce:2025}. \textit{Multi-agent} and \textit{adaptive dialogue systems} that consider user history, emotions, and culture could boost performance. Designing \textit{holistic evaluation frameworks} in real-world simulations to ensure privacy, explainability, fairness, and accessibility can help address AI risks~\cite{lawrence2024opportunities, yao2024survey}. At the policy level, NLP can analyze public sentiment on AI used for healthcare applications from actual users, and health guidelines to inform better regulations and health campaigns \cite{Lindquist:2021}. Importantly, \textit{interdisciplinary research} is needed to build AI-augmented therapeutic frameworks that complement human care—especially for underserved communities.

%% file: latex/education/education.tex
\section{\includegraphics[height=1.5em]{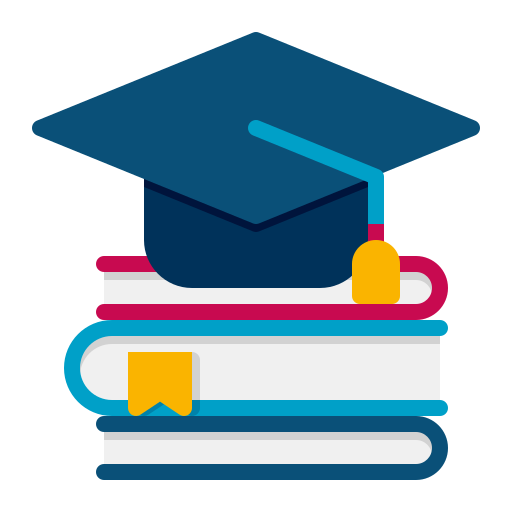} \hspace{0.1em}Education}

The integration of LLMs into education offers tools that support \textcolor{SDG4}{SDG4}: Quality Education, and also addresses a variety of societal risks by fostering informed and critical individuals (\textcolor{R4}{GR4}). NLP is used for automated feedback \cite{jurenka2024towards,bauer2023using,gao2024automatic,Stamper2024,ramesh2022automated}, tailored support \cite{kazemitabaar2024codeaid,daheim2024stepwise}, and self-paced learning \cite{kazemitabaar2023novices}. NLP tools can expand access in underserved regions \cite{yu2024whose}, bridge language gaps \cite{molina2024leveraging,kwak2024bridging}, assist learners with disabilities \cite{cheng2024llm}, and ease teacher workload \cite{lan2024teachers,wang-etal-2024-bridging,shridhar2022automatic}. As AI systems become more embedded in everyday life, AI literacy is essential. A recent review~\citep{yang2025navigating} draws connections with digital, data, and algorithmic literacies. However, NLP-driven efforts for AI literacy remain scattered~\citep{Long2020WhatIA, hr6791, moorkens-etal-2024-literacy, tapo-etal-2025-gaife, keris2025welcome}.

\paragraph{Challenges:} \textit{Model limitations}—such as lack of pedagogical reasoning~\cite{wang-demszky-2023-chatgpt, macina2023mathdial, Macina2025}, misaligned explainability~\cite{Okolo2024YouCB}, and accuracy~\cite{Stamper2024, kargupta-etal-2024-instruct} hinder the effective integration of NLP into education. \textit{Mixed perceptions and mistrust} toward AI also remain a barrier~\cite{Nader2022wm, Laupichler2024MedicalSA}. Broader issues like \textit{language, cultural differences, curriculum gaps, weak policy support, and limited infrastructure}—especially in developing regions—further restrict equitable access to AI/NLP in education~\cite{10.1145/3702163.3702449}.

\paragraph{Opportunities:} We believe LLMs can be a useful educational tool under sustained human oversight, but it cannot replace the relational, pedagogical, and contextual roles of human educators \citep{martynova-etal-2025-llms, 10.1093/jopedu/qhaf039}. Future work should focus on \textit{specific groups} such as teachers \cite{Du2024ExploringTE}, students \cite{Shen2024PerceivedSA}, and other professionals \cite{Lo2024EvaluatingAL} to obtain different perspectives on education. Moreover, \textit{aligning} with expert-annotated pedagogical traces and grounding evaluations in curriculum outcomes to bridge subject expertise and pedagogical effectiveness \cite{Macina2025, lucy-etal-2024-mathfish}. \textit{Human-in-the-loop} methods can scale expert strategies while preserving teacher agency, particularly in underserved settings \cite{Wang2024TutorCopilot}. \textit{Multi-agent simulations} provide privacy-preserving environments to test classroom policies and equity before real-world deployment \cite{zhang-etal-2025-simulating}, for instance, through Socratic planning agents that promote critical thinking over rote learning \cite{kargupta-etal-2024-instruct}. Finally, \textit{community-driven} efforts are key to consolidate best practices for safe and accountable educational AI \cite{Chu2025, Wen2024}.

%% file: latex/economy/economic_growth.tex
\section{\includegraphics[height=1.5em]{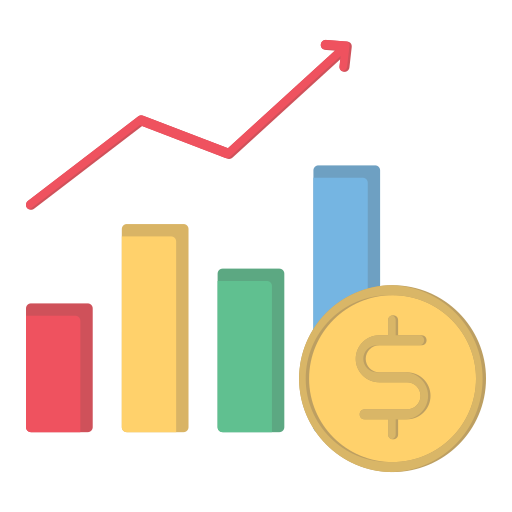} \hspace{0.1em}Poverty}
\label{sec:econ}

Economic downturn (\textcolor{R3}{GR3}) is the sixth highest-ranked GRs, with serious implications for poverty (\textcolor{SDG1}{SDG1})---one of the world’s most pressing challenges~\citep{lister2021poverty}. Nearly 700 million people continue to live on less than \$2.15 per day~\citep{international_poverty_line, people_in_extreme_poverty_count}. NLP methods have been used to extract socioeconomic patterns from news~\citep{lampos-etal-2014-extracting}, analyze text \citep{paterson2018representations, hoeschle2025let}, classify poverty status \cite{muneton2022classification}, extract poverty-related dimensions from interviews \citep{muneton2023identifying} or 
analyze global narratives around poverty \citep{curto-etal-2024-crime}, and to identify performance disparities among different socioeconomic groups \citep{cercas-curry-etal-2024-impoverished,nwatu-etal-2023-bridging}.

\paragraph{Challenges:} A major challenge is the fact that \textit{poverty data are often scarce or incomplete}~\cite{tingzon2019mapping, fatehkia2020relative}. Indicators like income or poverty status are rarely shared, making it hard to infer them from text. As a result, most socio-economic NLP studies rely on proxies such as mean income~\cite{hasanuzzaman2017temporal, abraham2020crowdsourcing}, education~\cite{cercas-curry-etal-2024-impoverished}, or (un)employment~\cite{preoctiuc2015studying, preoctiuc2015analysis}. These proxies may not accurately reflect true poverty levels and often vary across studies, resulting in \textit{limited comparability and impact}. 

\paragraph{Opportunities:} To enable a \textit{global analysis} of poverty, datasets labeled with income, socioeconomic status, or poverty indicators are essential, potentially gathered via data donations, user surveys, or social media statistics.\footnote{\scriptsize{\href{https://datareportal.com/social-media-users}{https://datareportal.com/social-media-users}}} Additionally, \textit{use-case-specific NLP applications} can advance poverty research~\citep{adauto-etal-2023-beyond}. For example, \textit{model analysis} can track the performance of current systems across socio-economic levels, \textit{information extraction} can monitor government funding for poverty alleviation, and \textit{machine translation} can improve resource access for non-English speakers. Finally, future work should develop \textit{clearer guidelines and taxonomies} for categorizing poverty-related data and NLP methods to ensure consistency and comparability across studies.

%% file: latex/peace_building_and_human_rights/peace.tex
\section{\reflectbox{\includegraphics[height=1.5em]{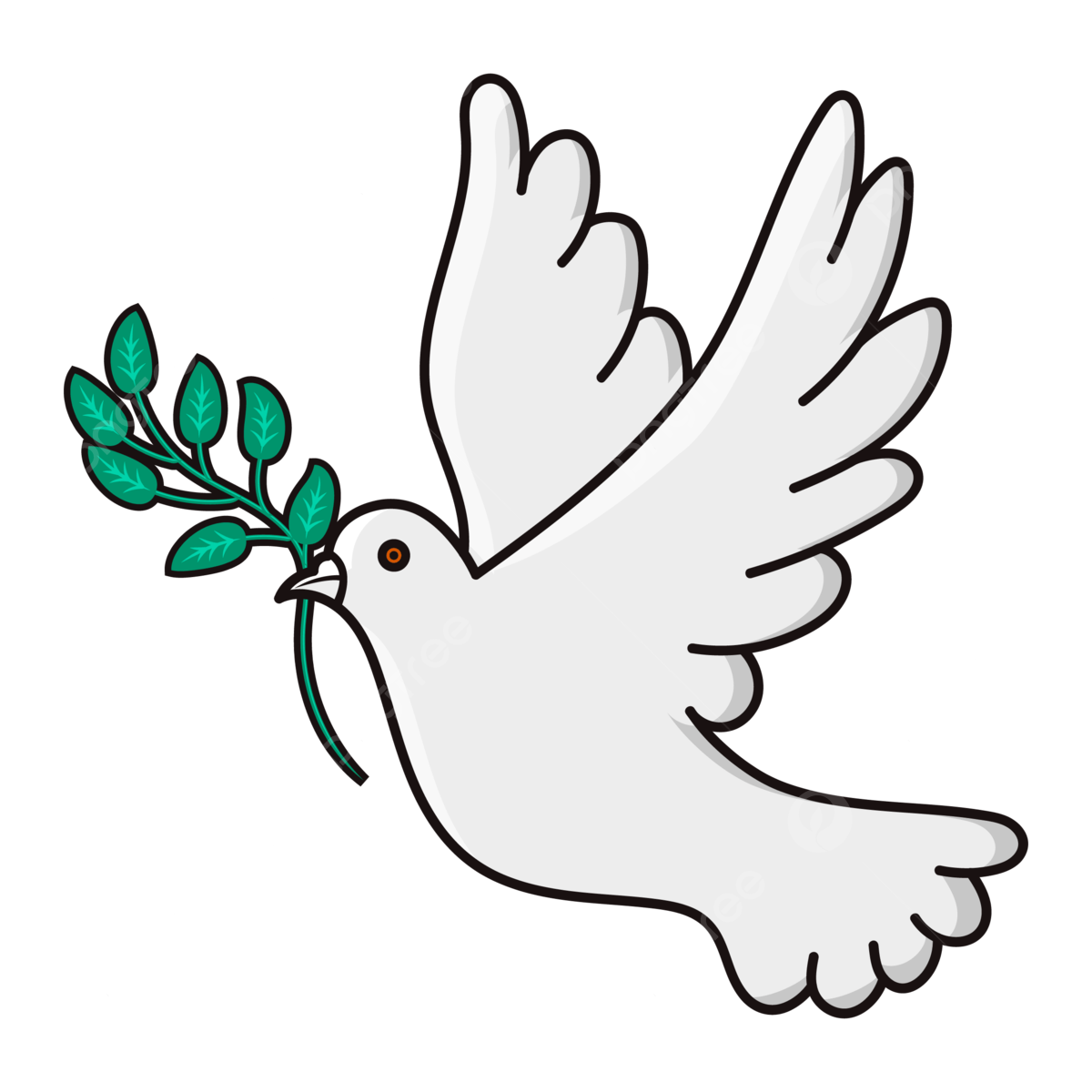}}\hspace{0.1em}Peacebuilding}
\label{sec:peace}

Peacebuilding is essential to achieving \textcolor{SDG16}{SDG16}, and despite escalating geopolitical threats (\textcolor{R1}{GR1}), NLP tools for peacebuilding like human rights monitoring, conflict prediction, and physical safety remains limited and underexplored.

\paragraph{\it Human Rights Violations.} NLP for human rights focuses on detecting violations across languages and platforms, from Arabic and Jordanian social media \citep{alhelbawy-etal-2016-towards, Khalafat_Alqatawna_Al-Sayyed_Eshtay_Kobbaey_2021} to Russian and Ukrainian Telegram \citep{nemkova2023detecting} and English Twitter \citep{pilankar-etal-2022-detecting}. Beyond general classification, efforts target threats to defenders \citep{ran-etal-2023-new}, forced labor with interpretable models \citep{guzman-etal-2024-towards}, human trafficking \citep{liu2023sweet, saxena2023idtraffickers, saxena2024matched}, dehumanization \citep{caporusso2024computational, burovova2024computational}, and other abuses \citep{yu2022again, guzman2024towards, sorato2024multilingual, de2024automated, wang2024metaphorical}.

\paragraph{\it Conflict Prediction.} NLP supports the extraction and forecasting of conflict dynamics \citep{halkia2020conflict, stoehr2021classifying, alsarra2023conflibert, sathvik2024ukrainian}. Methods span from topic modeling for early warning \citep{mueller2018reading, mueller2024introducing, Mueller_Rauh_Seimon_2024} to LLM-based approaches modeling escalation and resolution, including emerging agentic paradigms \citep{croicu2025from, nemkova2025large, nemkova2025agentic}. Studies also increasingly analyze civil unrest and protest dynamics \citep{sech2020civil, chinta2021study, scharf2021characterizing, raj2022cross, siskou2022automatized, wiedemann2022generalized, loerakker2024fine, olsen2024socio}. Advances include domain-adapted encoders like ConfliBERT for improved extraction and classification \citep{hu-etal-2022-conflibert} and automated situation reports via retrieval-augmentation (RAG) \citep{Nemkova2025TowardsAS}. Recent work extends to \textbf{diplomatic negotiation} and argument mining, analyzing discourse in peace talks and diplomacy \citep{glaser2022unsc, rodven2023unsc, zaczynska2024rhetorical, zaczynska2024diplomats, anisimova2024attitudes, poiaganova2025debates} by modeling pathways to resolution.

\paragraph{\it Physical Safety.} They have been used to analyze relevant social media \cite{ALGARADI2022-ipv, levy-etal-2022-safetext,Blandfort2019-gang, blevins-etal-2016-automatically}, police reports \cite{Karystianis2019-hv}, surveillance footage \cite{kumari2023-violence}, health records \cite{BORGER2022-psychiatric, Botelle2022-violence, macphaul2023-firearm}, reporting systems \cite{Chew2023-ma, arseniev2022-violent}, and news articles \cite{pavlick-etal-2016-gun}. Techniques such as topic modeling \cite{krishna2021-dv}, sentiment  analysis \cite{blevins-etal-2016-automatically}, entity extraction \cite{pavlick-etal-2016-gun}, and document classification \cite{chang-etal-2018-detecting} help identify patterns of violence, perpetrators, and victims, informing interventions and policy decisions.

\paragraph{Challenges:} Peacebuilding poses several challenges. \textit{Language and Context}: threats and violations are often euphemistic, coded, or strategically disguised, especially in authoritarian or conflict settings \cite{nemkova2023detecting, ran2023new, nemkova_llm2025}. Data is fragmented across low-resource languages and dialects, with annotation hindered by political sensitivity and expert disagreement \cite{Blandfort2019-gang, levy-etal-2022-safetext, Botelle2022-violence, Chew2023-ma}. \textit{Temporal Volatility}: conflicts evolve rapidly, as protests, ceasefires, or escalations outpace model adaptation, leaving systems trained on historical data prone to domain drift \cite{ALSaif2019, macphaul2023-firearm, parker2020-gunshot, chang-etal-2018-detecting}. \textit{High-Stakes Ethics}: misclassifications may expose activists, mislead humanitarian response, or legitimize violence, demanding stricter evaluation and oversight than other domains \cite{kumari2023-violence, ALGARADI2022-ipv, chang-etal-2018-detecting}. Since evidence for robust generalization remains limited, cautious interpretation and sustained human expertise are essential in deployment.

\paragraph{Opportunities:}
We believe that models such as conflict predictors and human rights violation detectors can support \textit{policymaking, interventions, and programmatic adjustments} across multilingual and cross-regional contexts \cite{ALSaif2019}.
NLP methods can also enhance \textit{operational efficiency}: LLMs enable rapid document review and synthesis for timely, context-aware decisions \cite{BORGER2022-psychiatric, Chew2023-ma}, while RAG systems improve access to relevant information \cite{Nemkova2025TowardsAS}. LLMs further assist with scenario generation, situational analysis, and strategic planning \cite{Nemkova2025TowardsAS}. Finally, a dedicated \textit{peacebuilding NLP workshop} could provide a crucial interdisciplinary forum for developing tools for humanitarian aid and crisis response.

%% file: latex/environment/climate.tex
\section{\includegraphics[height=1.5em]{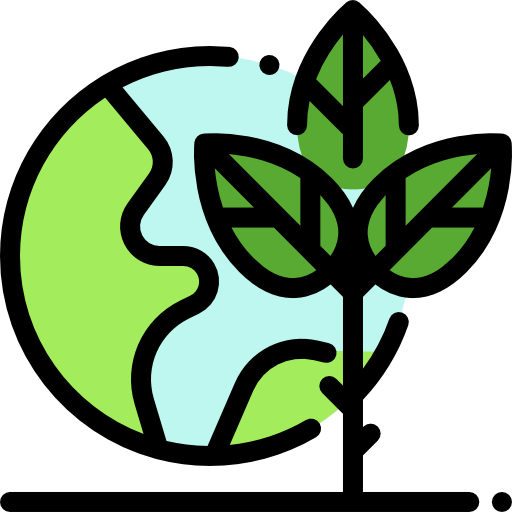}\hspace{0.1em}Environment Protection}\label{sec:climate}

NLP offers scalable tools for climate mitigation and adaptation—supporting key SDGs (\textcolor{SDG6}{SDG6}, \textcolor{SDG7}{SDG7}, \textcolor{SDG11}{SDG11}, \textcolor{SDG12}{SDG12}, \textcolor{SDG14}{SDG14}, \textcolor{SDG15}{SDG15}) and addressing critical global environmental risks (\textcolor{R2}{R2})—by extracting insights from unstructured text like scientific papers, policy reports, and assessments~\citep{IPCC2022}. Techniques such as topic modeling~\citep{sietsma-etal-2023-progress}, summarization~\citep{ghinassi-etal-2024-efficient}, and classification~\citep{varini-etal-2020-climatext, stammbach-etal-2023-environmental, bingler-etal-2022-cheaptalk, schimanski-etal-2023-climatebert-netzero} enable analysis of diverse datasets. NLP can also help to detect misinformation and greenwashing by verifying claims~\citep{diggelmann-etal-2020-climate-fever, hsu-etal-2024-evaluating}. While fine-tuning or pretraining LLMs on climate text is common~\citep{leippold-etal-2022-climatebert, thulke2024climategptaisynthesizinginterdisciplinary}, RAG-based chatbots~\citep{vaghefi-etal-2023-chatclimate} and automated fact-checkers~\citep{leippold-etal-2025-factchecking} are emerging rapidly.

\paragraph{Challenges:} Extracting quantitative information, particularly from sustainability reports, remains challenging due to \textit{the complex, multi-modal and non-standardized nature of the data sources}, leading to pipelines specifically designed for information extraction from tables~\citep{mishra-etal-2024-statements, dimmelmeier-etal-2024-informing}. Furthermore, LLMs, prone to \textit{hallucination}~\citep{vaghefi-etal-2023-chatclimate}, often output false, conflicting or outdated climate information \citep{fore-etal-2024-unlearning}.~\citet{bulian-etal-2024-assessing} evaluate the quality and factual accuracy of LLMs on climate information using a framework based on presentational and epistemological qualities.

\paragraph{Opportunities:}  
NLP can enable \textit{cross-disciplinary collaboration}, bridging domain experts from fields such as computer science, social science, and economics to analyze climate policies~\citep{gandhi-etal-2024-challenges} or bringing the scientific community and non-governmental organizations to uncover narratives in public climate discourse~\citep{Gehring2023, rowlands-etal-2024-predicting}. This is particularly important in expanding research to include \textit{cross-cultural and multilingual perspectives} \citep{zhou-etal-2024-large, bird-etal-2024-envisioning}. Lastly, NLP can inform other fields through climate reporting \citep{hershcovich-etal-2022-towards} and support \textit{sustainable behavior change} \citep{chockkalingam-etal-2025-go}.

%% file: latex/inequalities-inclusion/inequalities.tex
\section{\includegraphics[height=1.5em]{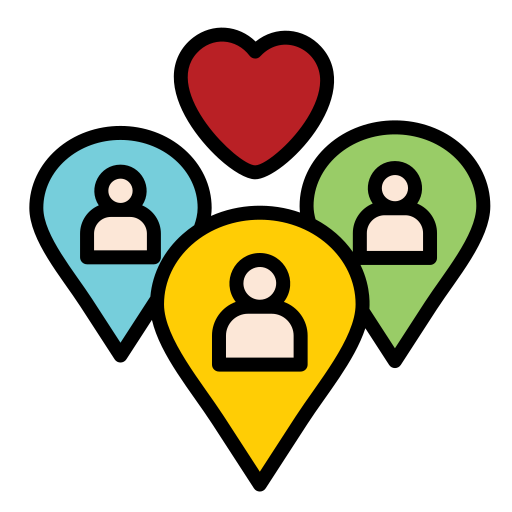}\hspace{0.1em}Inclusion and Inequalities}
\label{sec:ineq}

Inclusive and equitable language technologies are central to addressing systemic inequalities (disparities in how individuals or groups are represented, served, or affected), often reflecting socioeconomic, cultural, linguistic, and accessibility hierarchies. These efforts align with key sustainable development goals (\textcolor{SDG5}{SDG5}, \textcolor{SDG10}{SDG10}) and address inequality, ranked among the most severe societal risks in both the short and long term (\textcolor{R4}{GR4}).

A growing body of survey work examines bias and fairness in NLP, cataloguing demographic bias (gender, race, ethnicity, age, sexual orientation, disability, and socioeconomic status) \citep{gupta-etal-2024-sociodemographic}, its origins \citep{pmlr-v97-brunet19a, Hovy2021-ut}, alongside detection, quantification and mitigation methods ~\cite{stanczak2021survey,10.1145/3531146.3534627,10.1145/3700438,gallegos-etal-2024-bias}. Studies have focused on bias in model representations~\cite{bolukbasi2016man,caliskan2017semantics}, creating bias detection benchmarks \citep{zhao2018gender, parrish-etal-2022-bbq}, and testing models in high-stakes settings~\cite{de2019bias, Cross2024-yc}. More recent research addresses bias in agentic systems~\cite{borah-mihalcea-2024-towards}, proposing mitigation strategies including post-processing, interpretability-driven and causal approaches~\cite{attanasio2023tale,cai2024locating}, as well as counterfactual prompting~\cite{plyler2025iterative}.

Beyond demographic bias, inequalities also arise from the limited support of underrepresented communities with diverse accessibility needs~\cite{khanuja-etal-2023-evaluating}. NLP has been applied to a wide range of assistive technologies, including augmentative communication~\cite{park-etal-2022-pictalky}, text simplification~\cite{espinosa-zaragoza-etal-2023-automatic}, text-to-speech and speech recognition~\cite{kumar2023deep,li2022recent}, braille processing~\cite{tejesh2025multilingual}, image captioning and subtitling~\cite{stefanini2021show}, question answering~\cite{gurari2018vizwiz}, assistive chatbots~\cite{grassini2024systematic}, reading aids~\cite{wang2024gazeprompt}, and sign language translation~\cite{rust-etal-2024-towards}. 

\paragraph{Challenges:} Many challenges stem from simplifying assumptions, including binary gender representations, coarse socioeconomic proxies~\citep{bassignana-etal-2025-ai}, and unequal access to digital resources~\cite{10.1145/3600211.3604754}. Limited cross-linguistic and cross-cultural generalizability further constrains model reliability~\cite{stanovsky2019evaluating,adilazuarda-2024-beyond}. Culture—encompassing evolving norms, values, and worldviews—varies widely across communities~\citep{saha-etal-2025-meta}, yet NLP systems often reflect incomplete cultural perspectives, overrepresenting dominant linguistic narratives from high-resource languages and regions~\citep{hershcovich-etal-2022-challenges,karamolegkou-etal-2024-vision,mihalcea2025ai}.
Western-centric resources and the lack of intersectional frameworks reinforce marginalization~\cite{kleinberg2021algorithmic,sewunetie2024evaluating}, while compounded forms of discrimination—arising at the intersection of race, gender, social status, and disability—remain underexplored~\cite{stanczak2021survey,guo2021detecting,wald2021ai}. Progress is further hindered by scarce multimodal datasets, particularly for Braille and sign languages~\cite{hutchinson2020social,de-sisto-etal-2022-challenges,karamolegkou-etal-2025-evaluating}, limited interdisciplinary collaboration~\cite{kusters2020interdisciplinary}, and reliance on synthetic or prompt-generated data that obscures real-world biases~\cite{venkit2025study,morales2024gptbias}. Finally, opaque data sourcing practices limit accountability and reproducibility~\cite{bender2018data}.

\paragraph{Opportunities:} Inclusivity in NLP advances through participatory approaches centering marginalized voices, including co-designed fairness goals in gender bias mitigation~\cite{borah-mihalcea-2024-towards,ma-etal-2023-intersectional,lauscher-etal-2022-welcome} and collaborative data collection with cultural experts and underrepresented groups~\cite{unicef2020disability,hirmer2021building,bird2024centering,newman2024disability,karamolegkou-etal-2024-vision}. Solutions such as dynamic audit pipelines~\cite{park2023trak}, lightweight model editing~\cite{park2023trak,cai2024locating} and counterfactual data augmentation~\cite{zmigrod-etal-2019-counterfactual} help adapt models to sociocultural shifts. Fine-tuning LLMs can improve fairness and relevance~\cite{mai2024improving,bartl-leavy-2024-showgirls}. Future work should consider identity compositionality~\cite{welch-etal-2020-compositional} and pluralistic alignment reflecting complex social affiliations~\cite{10.5555/3692070.3693952}. Personalized, multimodal interaction design is vital for adaptive, accessible systems~\cite{paice2025assistive,wang2024gazeprompt}.

%% file: latex/technological_risks/digital_violence.tex
\section{\includegraphics[height=1.5em]{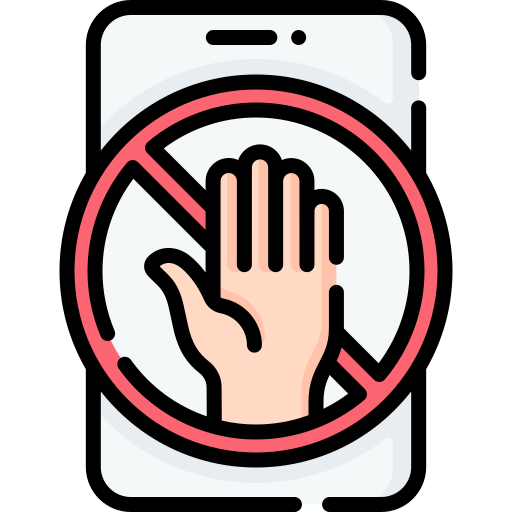} \hspace{0.1em}Digital Violence}
\label{sec:digital_violence}

The recent ease of access to digital devices (like smartphones and those based on IoT) has fueled the spread of digital violence globally~\citep{bjelajac2021specific}. This is central to Technological Risks, one of the most critical risks identified in the 2025 Global Risks Report (\textcolor{R5}{GR5}). There has been a large body of work on abusive/offensive/toxic/harmful speech classification~\cite{diazgarcia2025surveytextualcyberabuse}, generation of counter speech~\cite{bonaldi-etal-2024-nlp, saha-etal-2024-zero, wang-etal-2024-intent} and text detoxification~\cite{dementieva-etal-2025-multilingual, dale-etal-2021-text}, including several languages~\cite{aluru2020deeplearningmodelsmultilingual}.

\paragraph{Challenges:} Recent social media platform statements\footnote{\scriptsize{\href{https://about.fb.com/news/2025/01/meta-more-speech-fewer-mistakes}{https://about.fb.com/news/2025/01/meta-more-speech-fewer-mistakes}}} highlight the limitations of AI-based content moderation. Challenges include the \textit{subjective nature} of moderation, \textit{regional regulations}, and \textit{the culturally diverse} and implicit nature of content~\citep{ocampo-etal-2023-depth}. LLMs struggle with volatile topics without \textit{frequent fine-tuning}~\citep{roy-etal-2023-probing}, and efforts are hindered by \textit{unclear label taxonomies}, \textit{biases} in pre-trained models, and limited \textit{collaboration between lawmakers, platforms, and researchers}~\citep{yimam2024demarkedstrategyenhancedabusive}.

\paragraph{Opportunities:} 
We argue that future work should build low-resource, robust, and generalizable LLM moderation frameworks deployable in real time, requiring \textit{stronger collaboration between researchers, moderators, and policymakers}~\citep{Munzert2025-vy, bui-etal-2025-multi3hate}. A shared \textit{taxonomy of digital violence} is needed to reduce labeling ambiguity~\citep{yimam2024demarkedstrategyenhancedabusive}. Current text-centric moderation overlooks multimodal content such as video and audio, highlighting the need for \textit{multimodal LLMs}. To handle implicit and culturally diverse cases, future systems should integrate \textit{explainability}, \textit{fact-checking}, and real-time tools like RAGs and search APIs. Finally, moderation must move beyond hate speech to encompass the broader spectrum of digital violence—including harassment, bullying, spam, and abuse—through \textit{inclusive, context-aware systems} that reflect diverse cultural, regional, and linguistic realities~\cite{moghaddam2025towards, arora-etal-2023-probing}.

%% file: latex/misinformation/misinformation.tex
\section{\includegraphics[height=1.5em]{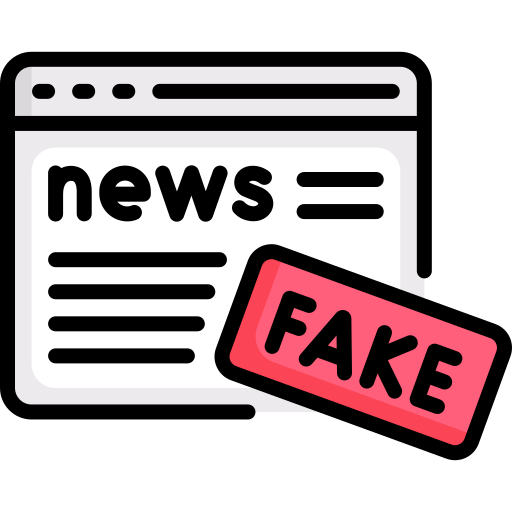}\hspace{0.1em}Misinformation}
\label{sec:mis}

Misinformation, the exposure to incorrect or misleading information, is ranked among the top five global risks in both short- and long-term scenarios (\textcolor{R5}{GR5}), and has a direct impact on the success of many of the SDGs.\footnote{\scriptsize{\href{https://www.un.org/sites/un2.un.org/files/information-integrity-and-sdgs-en.pdf}{https://www.un.org/sites/un2.un.org/files/information-integrity-and-sdgs-en.pdf}}} NLP tools can be both part of the solution, but also part of the problem (see \S\ref{sec:aiharms}). There are many approaches to tackle misinformation with tasks such as fake news detection, rumor classification, stance detection, and fact checking~\citep{oshikawa-etal-2020-survey, nakashole-mitchell-2014-language}. Early methods used stylometric features, while modern systems use neural models, pre-trained LMs, and techniques like RAG and prompt-based learning in multimodal and multilingual settings~\citep{akhtar-etal-2023-multimodal, mis2}. Recently LLM agents are used for generating fact-checking articles~\citep{sahnan2025llmsautomatefactcheckingarticle}, verifying complex claims~\citep{chowdhury-etal-2025-fact5, wang-etal-2025-piecing}, and handling long-form or machine-generated texts~\citep{xie-etal-2025-fire, boonsanong-etal-2025-facts}. 

\paragraph{Challenges:} There is a \textit{data scarcity}, especially in low-resource languages, emerging domains, and shifting distributions~\citep{guo-etal-2022-survey}. \textit{Multilingual and conflicting evidence} complicates verification~\citep{schlichtkrull-2024-generating, zhang-etal-2024-need}, while \textit{low-visibility claims} targeting marginalized groups often go unchecked~\citep{guo-etal-2022-survey}. In \textit{high-stakes domains} like healthcare and politics, reliable detection is critical to prevent real-world harm~\citep{abdul-mageed-etal-2021-mega, zhao-etal-2023-panacea}, yet systems still lack \textit{robustness, fairness, and explainability}. Lastly, LLMs pose \textit{misuse risks} by generating convincing falsehoods~\citep{buchanan2021, gabriel-etal-2024-misinfoeval}.

\paragraph{Opportunities:} A key opportunity lies in developing \textit{human-centered evaluation} methods tailored to real-world tasks~\citep{DAS2023103219}, and strengthened through \textit{interdisciplinary collaboration}--integrating social and economic theories~\citep{mis4} and partnerships with moderators, policymakers, and fact-checkers~\citep{10.1145/3706598.3713277}. Improving the \textit{timeliness of ground truth} via LLM-agents that access web evidence and external knowledge bases can enhance adaptability to fast-evolving content. Developing \textit{domain-agnostic features} generalizable across topics, languages, and modalities can help detect shifting deceptive styles~\citep{mis2} and identify check-worthy claims. The growing threat of LLM-generated misinformation remains a critical frontier~\citep{liu-etal-2025-survey, gabriel-etal-2024-misinfoeval}, underscoring the need for adaptive, accountable, and transparent detection systems.

%% file: latex/technological_risks/harms.tex
\section{\includegraphics[height=1.5em]{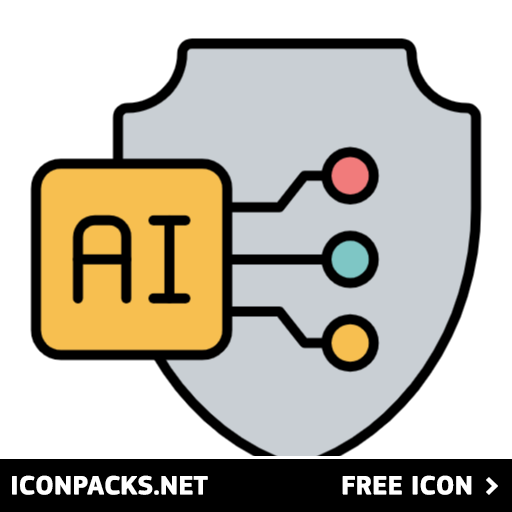}\hspace{0.1em}AI Harms}
\label{sec:aiharms}

AI harms, referring to negative impacts stemming from the design, deployment, or use of AI systems, are among the top global threats (\textcolor{R5}{GR5}). NLP is part of the problem—but also part of the solution. We first consider the black-box nature of AI systems and then follow the harm taxonomy by~\citet{aiharmstaxo} to structure our discussion and explore NLP-based mitigation strategies.

\paragraph{\it The Black Box Problem.} LLMs often operate as black boxes, offering limited interpretability and transparency \citep{hassija2024interpreting}. With little insight into their design or training, the public and AI community remain reliant on what creators choose to disclose. To address this, NLP research has explored textual explanations, such as summaries \cite{atanasova-etal-2020-generating-fact,kotonya-toni-2020-explainable-automated}, contrastive \cite{schuster-etal-2021-get}, and counterfactual forms \cite{yang-etal-2020-generating,tolkachev-etal-2022-counterfactual}. Visual methods like LIME \citep{ribeiro-etal-2016-trust}, ACE \citep{ghorbani2019towards}, and heatmaps \citep{interpret} highlight input relevance, while structural tools like SVCCA \citep{NIPS2017_dc6a7e65}, t-SNE, and TCAV \citep{pmlr-v80-kim18d} uncover concept-level insights.

\paragraph{\it Representation and Toxicity.} NLP techniques are actively developed to address biases in training data and AI models~\citep{gallegos-etal-2024-bias}, to reduce bias and toxicity. Further discussion is provided in~\S\ref{sec:ineq} on inequality and~\S\ref{sec:digital_violence} on digital toxicity.

\paragraph{\it Privacy, Safety and Malicous Uses.}
NLP approaches to harm reduction include learning to refuse or prevent memorization to protect personal data~\citep{carlini_data, liu-etal-2025-learning}, adversarial training~\citep{10_adversarial}, safe decoding~\citep{xu-etal-2024-safedecoding}, authorship verification~\citep{huang-etal-2024-large}, safety alignment~\citep{bhardwaj-etal-2024-language}, and red-teaming to expose vulnerabilities~\citep{purpura-etal-2025-building}.

\paragraph{\it Ungrounded Knowledge.} As pointed in \S\ref{sec:mis} LLM outputs can include hallucinations and spread misinformation~\citep{pan-etal-2023-risk, sun2024trustllm, mis2}. Mitigation strategies include grounding with external knowledge via RAG~\citep{lewis2020retrieval, peng2023check, asai2023self}, expressing uncertainty~\citep{yang2023alignment, feng-etal-2024-dont, xiong2024llmsexpressuncertaintyempirical, deng-etal-2024-dont}, and self-verification methods~\citep{weng-etal-2023-large, hong-etal-2024-closer}.

\paragraph{\it Socioeconomic and Environmental Harms.}
In NLP, researchers have called for greater climate awareness and transparency through reporting frameworks \citep{hershcovich-etal-2022-towards}, while others focus on reducing energy use with model optimization techniques like pruning, quantization, and distillation \citep{schwartz2020green, jin-etal-2024-comprehensive, zhu-etal-2024-survey-model}. Tackling socioeconomic harms (\S\ref{sec:econ}, \S\ref{sec:ineq}) further requires closer collaboration with sociology and HCI \citep{hcinlp-ws-2024-bridging, nlpcss2024} to further enhance the limited work in this direction.

\paragraph{Challenges:} Key barriers to prevent AI harms include transitioning from prototype to deployment which entails coping with \textit{data distributional shifts} between ``lab'' and ``field'' data, and managing \textit{bias or subjective labels}. Many current benchmarks often overlook \textit{low-resource} or politically \textit{sensitive domains}~\citep{kim-etal-2025-representing, cho-etal-2025-hermit}.
Additionally, the lack of standardized practices for measuring \textit{real-world impact} complicates evaluation beyond standard metrics. The \textit{effectiveness} of the proposed tasks is often questioned ~\citep{chen2024llmgenerated, kotonya-toni-2024-towards}, and there are very \textit{few rigorous and holistic evaluation frameworks} \cite{atanasova-etal-2023-faithfulness, liang2023holistic}. 
There are also several governance problems and establishing \textit{equitable partnerships} with nonprofits, where power imbalances or misaligned goals can hinder collaboration. Lastly, while NLP methods can mitigate certain AI harms, many of them—such as overreliance or erosion of human agency—cannot be resolved through technical solutions alone.

\paragraph{Opportunities:} \textit{Field trials} of NLP systems under real-world conditions are necessary, especially in low-resource or use-case-specific settings, while encouraging participatory design and critical user engagement~\citep{10.1145/3708359.3712112}. On the modeling side, advances in \textit{retrieval-augmented generation (RAG)} and \textit{knowledge augmentation} methods can improve reliability by enhancing credibility~\citep{xu-etal-2024-knowledge-conflicts, chen-etal-2024-seer, ChenDT25}. \textit{Multidisciplinary approaches} such as logical relationships between inputs~\citep{AyoobiPT25, FreedmanDG00T25} and mechanistic interpretations of model behavior~\citep{hou-etal-2023-towards, yu-ananiadou-2024-interpreting} can further enhance transparency. Additional studies could examine LLM overreliance, manipulation, the uneven distribution of benefits from model access, as well as tracing gaps and overlaps between professional and laypeople AI concerns~\citep{karamolegkou-etal-2025-ethical}. Embracing \textit{process-aware NLP} enables alignment with domain logic and ethical goals through explainability and fairness-by-design \citep{Bernardi2024BPLLM, zhuang2025docpuzzleprocessawarebenchmarkevaluating}. Finally, \textit{mapping existing frameworks} for AI4SG and NLP4PI~\citep{FATML:2020:AccountableAlgorithms, floridi2021design}, suggested actions \cite{NIST2024AI6001} and principles for trustworthy AI \citep{oecd}, can support the development of adaptable, unified guidelines for responsible NLP deployment.

%% file: latex/3_summary.tex
\section{Summary and Call to Action}

To achieve the full potential of NLP4SG, the community should move beyond task-centric innovation toward impact-driven systems that are safe, inclusive, and globally relevant. This work reviewed nine key application areas, addressing \textbf{RQ1} on existing impactful solutions with concrete examples of how modern NLP technologies have already contributed. Complementing this review, our quantitative analysis of 47K ACL papers showed that research on AI harms and inclusion has grown most rapidly, while domains like poverty, peacebuilding, and environment remain underexplored (Figure \ref{fig:overview} and Appendix \ref{sec:appendix-llm-annotation}). At the same time, in addressing \textbf{RQ2} on key open challenges, we find that areas like misinformation, online harms, and education have seen sustained attention in NLP, while domains such as poverty alleviation, environmental protection, and peacebuilding are only starting to gain traction in response to real-world crises. Addressing \textbf{RQ3} concerning promising directions for future work, we outline a set of universal challenges, opportunities, and actionable recommendations to guide future research in NLP4SG:

\textbf{Emerging Challenges and Opportunities:}
Despite barriers such as \textit{data scarcity and representational bias}, \textit{misaligned evaluation metrics}, and persistent \textit{safety, privacy, and ethical concerns}, alongside ongoing \textit{infrastructural gaps}, we identify several promising directions for progress.
(1) \textit{Multilingual and multicultural learning} can help systems better reflect real-world diversity including not only mainstream languages, but caring about local language communities equally; 
(2) \textit{Human–AI collaboration} enables more adaptive and interpretable NLP pipelines; (3) \textit{Participatory design and evaluation} ensure that systems are co-developed with affected communities; (4) \textit{Retrieval-augmented and policy-aware methods} provide tools for verifiable, context-sensitive applications; and (5) \textit{Explainability and AI literacy} foster critical engagement and equitable access.

\textbf{Call to Action:} To advance NLP4SG, we call on the community to: 
(1) Develop joint benchmarks featuring \textbf{multilingual}, \textbf{culturally diverse}, and \textbf{socially grounded} data;
(2) Collaborate closely with \textbf{domain experts}, such as educators, health practitioners, and civil society organizations, to co-design evaluation frameworks that reflect end-user needs.
(3) Pursue \textbf{human-centered methodologies} instead of one-size-fits-all solutions. Progress depends on pluralistic, context-aware roadmaps that align with both local realities and global development goals.
(4) Finally, while modern LLMs offer significant potential,
it is crucial to ensure their \textbf{affordability} and \textbf{accessibility} so they serve the public good rather than exacerbate existing inequalities. Assess whether deploying a large, one-size-fits-all generalist LLM is truly necessary, or whether \textbf{more efficient}, and \textbf{more environmentally sustainable} NLP solutions would be preferable.

NLP has the tools to move beyond abstract benchmarking and toward socially responsive technologies designed with---and for---impacted communities. Realizing this vision requires not just technical innovation but also sustained interdisciplinary collaboration, inclusive practices, and a commitment to long-term global equity. We hope our findings can help researchers early in their  careers to find their research niche and that more advanced researchers will have a fresh overview of the field to foster NLP4SG applications with a more interdisciplinary paradigm.

%% file: latex/4_Acknowledgments.tex
This work was supported by the Novo Nordisk Foundation (grant NNF20SA0066568), which funded the research of Anders Søgaard and Antonia Karamolegkou. Antonia Karamolegkou was also supported by the Onassis Foundation - Scholarship
ID: F ZP 017-2/2022-2023.
Zhijing Jin was supported partly by the German Federal Ministry of Education and Research (BMBF): Tübingen AI Center, FKZ: 01IS18039B; by the Machine Learning Cluster of Excellence, EXC number 2064/1 – Project number 390727645; and by Schmidt Sciences SAFE-AI Grant. Daryna Dementieva's and Alexander Fraser's work was supported by the European Research Council (ERC) under grant agreement No. 101113091 - Data4ML, an ERC Proof of Concept Grant as well as co-funded by the an ERC Advanced Grant EPICAL, grant agreement No. 101141712. Views and opinions expressed are however those of the author(s) only and do not necessarily reflect those of the European Union or the European Research Council. Neither the European Union nor the granting authority can be held responsible for them. Daryna Dementieva's work was additionally supported by the Friedrich Schiedel TUM Think Tank Fellowship.